\def\year{2021}\relax
\documentclass[letterpaper]{article} % DO NOT CHANGE THIS
\usepackage{aaai21}  % DO NOT CHANGE THIS
\usepackage{times}  % DO NOT CHANGE THIS
\usepackage{helvet} % DO NOT CHANGE THIS
\usepackage{courier}  % DO NOT CHANGE THIS
\usepackage[hyphens]{url}  % DO NOT CHANGE THIS
\usepackage{graphicx} % DO NOT CHANGE THIS
\usepackage{graphicx}  % DO NOT CHANGE THIS
\graphicspath{{FIGURES/}}
\usepackage{natbib}  % DO NOT CHANGE THIS AND DO NOT ADD ANY OPTIONS TO IT
\usepackage{caption} % DO NOT CHANGE THIS AND DO NOT ADD ANY OPTIONS TO IT
\usepackage{amsmath}
\usepackage{amssymb}
\usepackage{booktabs}
\usepackage{diagbox}
\usepackage[switch]{lineno}
\title{Fooling Thermal Infrared Pedestrian Detectors in Real World Using Small Bulbs}
\author{
    Xiaopei Zhu\textsuperscript{\rm 1},
    Xiao Li\textsuperscript{\rm 2},
    Jianmin Li\textsuperscript{\rm 2},
    Zheyao Wang\textsuperscript{\rm 1}\thanks{Corresponding Author},
    Xiaolin Hu\textsuperscript{\rm 2*}
}
\affiliations{
    \textsuperscript{\rm 1}Department of Microelectronics and Nanoelectronics, Tsinghua University, Beijing 100084, China\\
    \textsuperscript{\rm 2}Department of Computer Science and Technology, Institute for Artificial Intelligence\\
    State Key Laboratory of Intelligent Technology and Systems, BNRist, Tsinghua University, Beijing 100084, China\\
    \{zxp18, lixiao20\}@mails.tsinghua.edu.cn,
    \{lijianmin, z.wang, xlhu\}@tsinghua.edu.cn

}
\begin{document}
\maketitle
% \linenumbers
\begin{abstract}
Thermal infrared detection systems play an important role in many areas such as night security, autonomous driving, and body 
temperature detection. They have the unique advantages of passive imaging, temperature sensitivity and penetration. 
But the security of these systems themselves has not been fully explored, which poses risks in applying these 
systems. 
% We propose a physical board attack method against the state-of-the-art pedestrian detectors. 
We propose a physical attack method with small bulbs on a board against the state of-the-art pedestrian detectors.
Our goal is to make infrared pedestrian detectors unable to detect real-world pedestrians. 
Towards this goal, we first showed that it is possible to use two kinds of patches to attack the infrared 
pedestrian detector based on YOLOv3. The average precision (AP) dropped by 64.12\% in the digital world, while a blank board with the same size caused the AP to drop by 29.69\% only. 
After that, we designed and manufactured a physical board and successfully attacked 
YOLOv3 in the real world. In recorded videos, the physical board caused AP of the target detector to drop by 34.48\%, %Besides, we improved the transferbility of the attack on other detectors by using model ensemble.
 while a blank board with the same size caused the AP to drop by 14.91\% only.
With the ensemble attack techniques, the designed physical board had good transferability to unseen detectors.  
% As far as we know, we are the first to realize physical attacks on the thermal infrared pedestrian detector.
\end{abstract}

\section{Introduction}
% \noindent Infrared object detection has been widely used in many areas. Infrared object 
% detection has its unique advantages. First of all, the infrared object detection system can work at night, 
% and some autonomous driving companies are currently considering using it as an auxiliary system at night. 
% Secondly, the infrared detection system can detect objects through some obstacles. For example, a person can 
% still be detected when hiding in bushes. Besides, compared to visible light images, infrared images not only 
% contain the shape information of the object but also contain the temperature information of the object. With 
% the development of deep learning, infrared object detection has made significant progress. During COVID-19, 
% infrared pedestrian detection was widely used.
\noindent Deep learning has achieved remarkable success in various tasks such as classification \cite{DBLP:conf/cvpr/KarpathyTSLSF14}, detection \cite{DBLP:conf/cvpr/RedmonF17}, and segmentation \cite{DBLP:journals/Cognitive/ZiYin}. 
However, it is well known that deep neural networks (DNNs) are vulnerable to adversarial attacks, i.e., they can be fooled by input examples with some deliberately designed small perturbations. Such examples are called adversarial examples. Since the findings of Szegedy et al.
(2013)\nocite{DBLP:journals/corr/SzegedyZSBEGF13}, there is increasing interest in the field of adversarial 
attacks. For digital world attacks, many methods have been proposed including the gradient-based  attacks \cite{DBLP:journals/corr/GoodfellowSS14,
DBLP:conf/iclr/KurakinGB17,DBLP:conf/iclr/MadryMSTV18}, optimization-based attacks \cite{DBLP:conf/sp/Carlini017,
DBLP:journals/corr/SzegedyZSBEGF13,DBLP:conf/cvpr/EykholtEF0RXPKS18}, and network-based attacks \cite{DBLP:conf/ijcai/XiaoLZHLS18,DBLP:conf/aaai/LiuLFMZXT19}.
Annoyingly, adversarial examples not only exist in the digital world but also exist in the real world. 
Athalye et al. (2018)\nocite{DBLP:conf/icml/AthalyeEIK18} showed that a 3D printed turtle could be mistaken for a rifle by a DNN. Simen et al. 
(2019)\nocite{DBLP:conf/cvpr/ThysRG19} designed a printable patch that successfully attacked the pedestrian detection system and achieved an ``invisibility''
effect. Xu et al. (2019)\nocite{DBLP:journals/corr/abs-1910-11099} invented Adversarial T-shirts, which could attack the person detectors even if it has 
non-rigid deformation due to a moving person's pose changes. The adversarial attack in the physical world has attracted much attention as it poses high risks to the widely deployed deep learning-based security systems.
It urges researchers to re-evaluate the safety and reliability of these systems. 
\begin{figure}[tb]
\centering
\includegraphics[width=0.8\columnwidth]{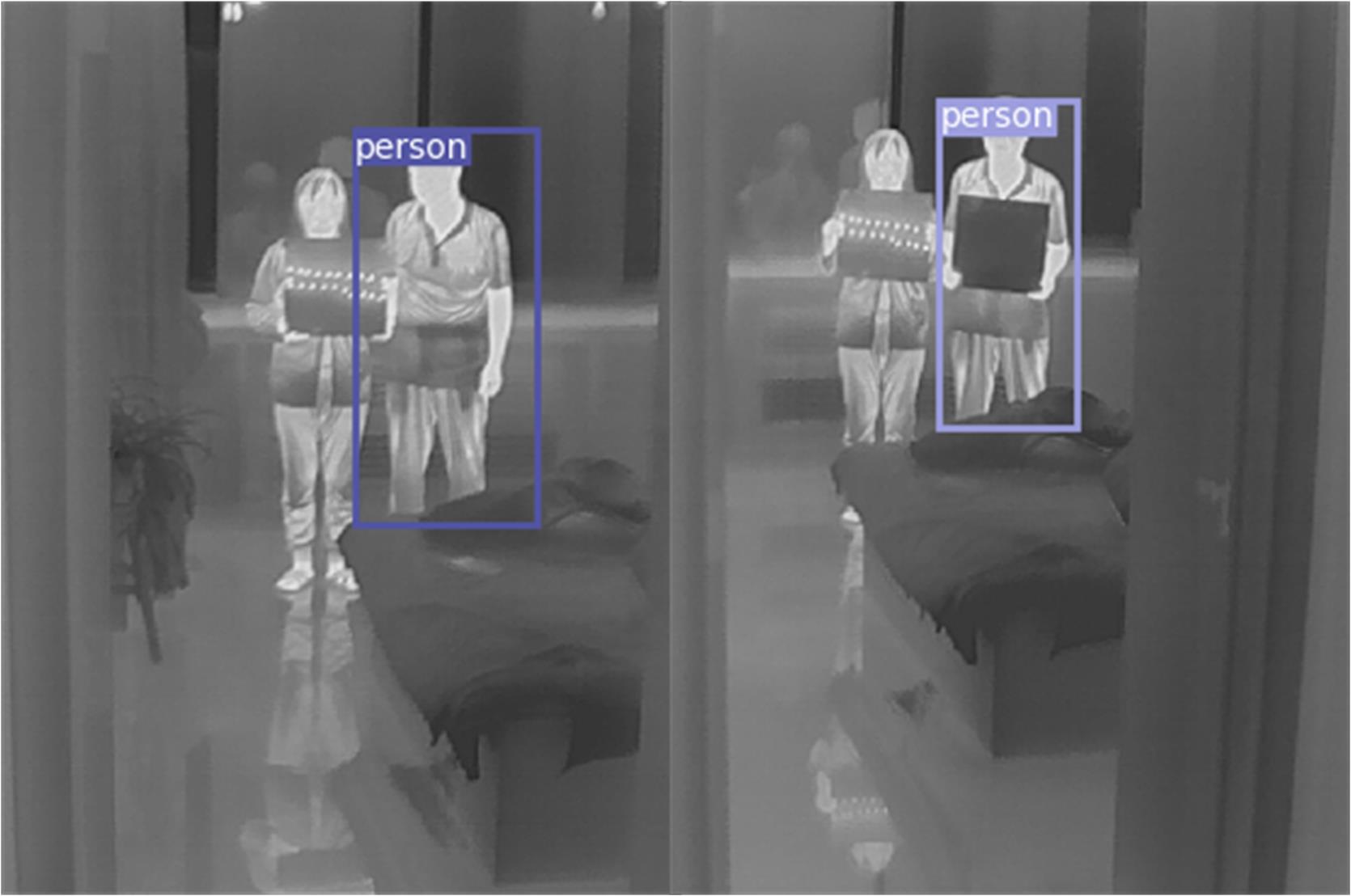} % Reduce the figure size so that it is slightly narrower than the column. Don't use precise values for figure width.This setup will avoid overfull boxes. 
\caption{An example of physical infrared attack and control experiments}
\label{fig1}
\end{figure}

% However, there is still a lack of research on the safety of infrared object detection systems. In this 
% paper, we try to focus on the security of infrared detection. In contrast, there is much work on adversarial 
% attacks in the visible light field. Since the findings of Szegedy et al.(2014)\nocite{DBLP:journals/corr/SzegedyZSBEGF13}
% , there is increasing interest in the field of adversarial attacks. Adversarial examples refer to input samples 
% formed by deliberately adding interference to the data set, which will cause the model to give an incorrect output 
% with high confidence. According to the access rights to the model, there are white-box attacks and black-box attacks.
% In the white box attack, we can obtain all the parameters of the network, but in the black box attack, we can only 
% get the input and output of the network. According to the attack targets, there are targeted attacks and non-targeted attacks.
% Take classification attack as an example. Targeted attacks mean that the defense model classifies the attack image into the 
% target category, while non-targeted attacks mean that the defense model classifies the attack image into any 
% category other than the correct category. According to different attack domains, there are digital attacks and physical attacks. 
% Digital attacks assume that the attacker can directly input the disturbed digital image into the DNN network, and 
% physical attacks study the attack process of adversarial examples in the physical world on DNN.

Almost all current research on adversarial attacks focused on the visible light field. There is a lack of research 
on the safety of infrared (Thermal infrared in our paper) object detection systems, which have been widely deployed in our society with their unique 
advantages. First of all, the infrared object detection systems can work at night. It implies that the 
surveillance systems based on the infrared cameras do not need environmental light and can save energy 
in certain scenarios. Some autonomous driving companies are currently using infrared images as auxiliary 
input at night. Secondly, they can detect objects behind certain obstacles. For example, a person can still 
be detected when hiding in bushes. Thirdly, compared to visible light images, infrared images not only contain the shape information of the object 
but also contain the temperature information of the object. During the pandemic caused by COVID-19, infrared pedestrian detection has received more and more attention. With the development of deep learning, infrared 
object detection has made significant progress.

Compared with the visible images which have three channels (RGB), the challenge of the infrared image processing is that the infrared image has only one gray-scale channel, and the texture information is far less than that of 
visible light image. Besides, to realize physical attacks, visible images can be printed by a laser printer, which can preserve most details of the designed adversarial images. Obviously, one cannot obtain an adversarial infrared image by ``printing'' any digital image.

% However, thermal infrared images are not easy to ``print" in the physical world. 
To solve this problem, we propose a method to realize the adversarial infrared images in the real world.
%In this paper, our contribution is that we design and manufacture a physical board that could attack the infrared pedestrian detection system in real world, and we can adjust its pattern for different situations.
Our method is to use a set of small bulbs on a cardboard, which can be held by hands. A dedicated circuit is designed 
to supply electricity to the bulbs. Eqipped with eight 4.5V batteries, the cardboard decorated with small bulbs 
can successfully fool the state-of-the-art infrared pedestrian detection models. The cost of this setup is less 
than 5 US dollars. An example of physical infrared attack and control experiments is shown in Figure \ref{fig1}. 
As far as we know, we are the first to realize physical attacks on the thermal infrared pedestrian detector.
% The overview of the process is shown in Figure \ref{main_process}.

% \begin{itemize}
% \item We propose a patch attack method that can effectively reduce the accuracy of infrared pedestrian recognition systems in the real world.
% \item We design two methods for generating patches for infrared object attacks, including pixel-level patch and Gaussian functions patch.
% \item We design and manufacture a patch that could attack the infrared pedestrian detection system physically, and we can adjust its pattern for different situations.
% \item We demonstrate that the physical board we designed could still attack the infrared object detection systems after scale transformation and pattern fine-tuning.
% \end{itemize}

\section{Related Work}

% \noindent In this paper, we attack the object detection model both in the digital world and in the physical world. Therefore, in this section, we investigate the related work of digital world attacks and physical world attacks, as well as the popular object detection model.

\subsection{Digital Attacks in Visible Light Images}

Szegedy et al. (2014) found that when small perturbations added to an image, it will cause image 
classification errors. Goodfellow et al. (2015)\nocite{DBLP:journals/corr/GoodfellowSS14} developed a method to efficiently compute an adversarial 
perturbation for a given image, which is called FGSM. Based on the above work, BIM \cite{DBLP:conf/iclr/KurakinGB17} optimizes the process by 
taking more small-step iterations. Dong et al. (2018)\nocite{DBLP:conf/cvpr/DongLPS0HL18} proposed MIFGSM, which added a momentum term during 
iterative optimization. A particular case of digital attacks is to modify only one pixel of the image to 
fool the classifier. Moosavi-Dezfooli et al. (2017)\nocite{DBLP:conf/cvpr/Moosavi-Dezfooli17} computed universal perturbations to fool DNNs.

\begin{figure*}[htbp]
\centering
\includegraphics[scale=0.40]{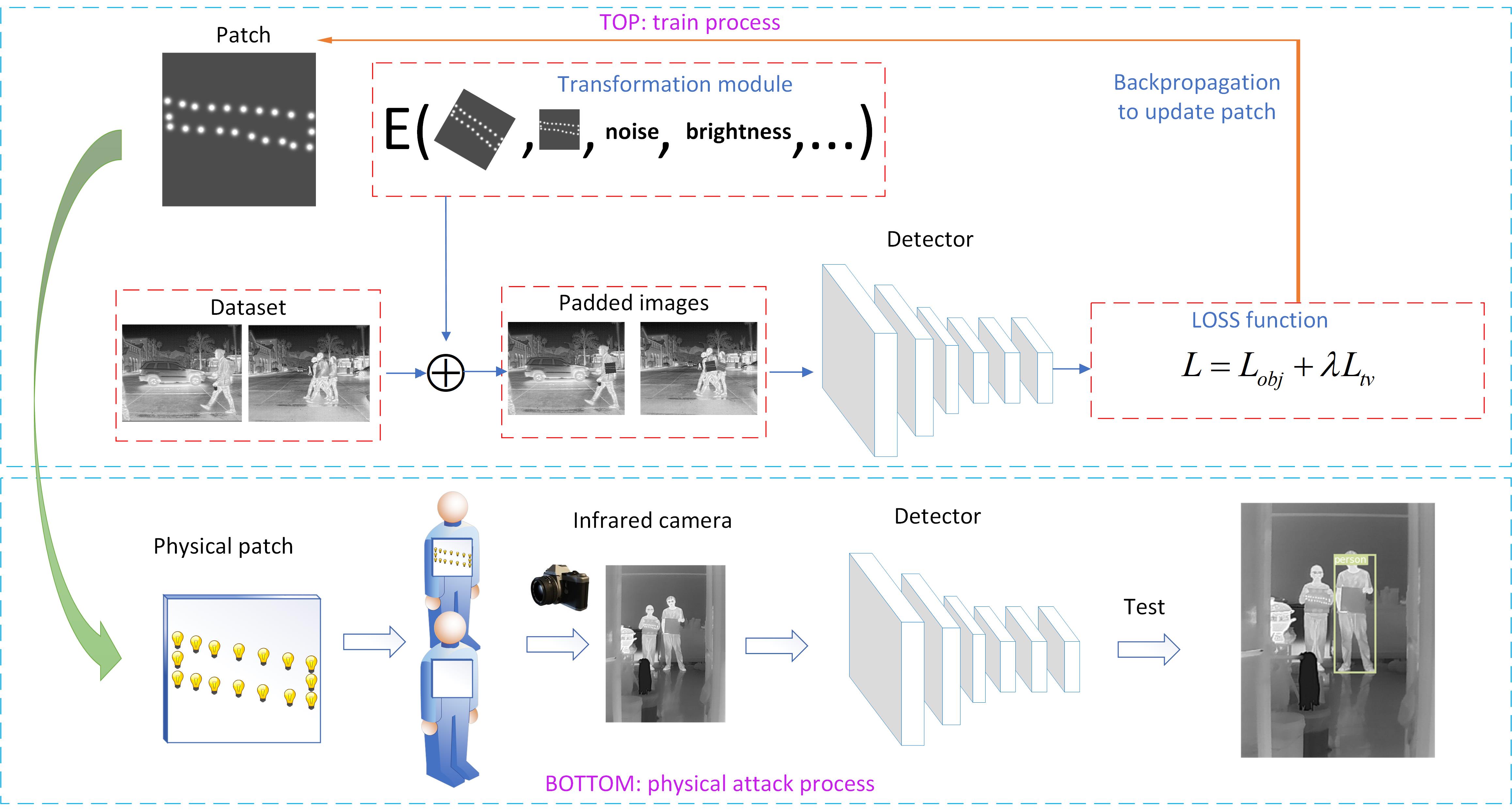} % Reduce the figure size so that it is slightly narrower than the column. Don't use precise values for figure width.This setup will avoid overfull boxes. 
\caption{Overview of our work. Top is the training process, bottom is the physical attack process.}
\label{main_process}
\end{figure*}

\subsection{Physical Attacks in Visible Light Images}
Sharif et al. (2016)\nocite{DBLP:conf/ccs/SharifBBR16} designed a wearable glasses that could attack facial 
recognition systems. Eykholt et al. (2018)\nocite{DBLP:conf/cvpr/EykholtEF0RXPKS18} designed a road sign with perturbations that can fool the road sign 
recognition systems in practical driving. Athalye et al. (2018)\nocite{DBLP:conf/icml/AthalyeEIK18} proposed a method for generating 3D adversarial 
examples. Zhou et al. (2018)\nocite{DBLP:journals/corr/abs-1803-04683} designed an invisible mask that can attack the face recognition system. 
They hide the infrared diode under the hat. Simen et al. (2019)\nocite{DBLP:conf/cvpr/ThysRG19} proposed an optimization-based method to create 
a patch that could successfully hide a person from a person detector.

\subsection{Object Detection}
% \noindent 
% In this paper, we mainly attacked the YOLOv3 model\cite{DBLP:journals/corr/abs-1804-02767}. 
% There are three types of target detectors: one-stage, two-stage, and multi-stage. The YOLOv3\cite{DBLP:journals/corr/abs-1804-02767} 
% model is a typical one-stage detection method. Its advantage lies in the ability to maintain a 
% faster detection speed while maintaining higher detection accuracy. 
% % Compared with the previous generation YOLOv2\cite{DBLP:conf/cvpr/RedmonF17}, YOLOv3 has a significant improvement in the detection performance of small targets. 
% RetinaNet is also a one-stage detector using Focal loss. 
% % Focal Loss adds a factor to the original cross-entropy loss function to make the loss function pay more attention to hard examples.Faster RCNN is a typical two-stage detector. It solves the problem that Fast R-CNN still needs to extract region proposal in advance, resulting in low time efficiency.
% Faster R-CNN is a typical two-stage detector which designed RPN (Region Proposal Networks) to extract candidate ROI regions.
% % , and RPN results are then fed detection network to perform target detection. 
% Cascade R-CNN is a multi-stage detector. The model contains a set of detectors trained with increasing IOU 
% thresholds. 
There are three kinds of target detectors: one-stage, two-stage, and multi-stage detectors. The YOLOv3 \cite{DBLP:journals/corr/abs-1804-02767} model is 
a typical one-stage detection method with the advantage of fast detection speed and high accuracy. RetinaNet \cite{DBLP:conf/iccv/LinGGHD17} is 
also a one-stage detector using Focal loss. Faster R-CNN \cite{DBLP:journals/pami/RenHG017} is a typical two-stage detector using RPN to extract 
candidate ROI regions. Cascade R-CNN \cite{DBLP:conf/cvpr/CaiV18} is a multi-stage detector which contains a set of detectors trained with 
increasing IOU thresholds.
Biswas et al. (2017)\nocite{DBLP:journals/tip/BiswasM17} used local steering kernel (LSK) as low-level descriptors for detecting pedestrians in 
thermal infrared images. Zhao et al. (2019)\nocite{DBLP:conf/apsipa/ZhaoCZZP19} proposed an infrared pedestrian detection method with a converted 
temperature map. YuHang et al. (2020)\nocite{DBLP:journals/access/JiangDCX20} proposed an infrared small target detection algorithm based on the peak 
aggregation number and Gaussian discrimination. 
Mate et al. (2020) investigated the task of automatic person detection in thermal images
using convolutional neural network models originally intended for detection in RGB images. They compared the
performance of the standard state-of-the-art object detectors such as Faster R-CNN, SSD, Cascade R-CNN,
and YOLOv3, that were retrained on a dataset of thermal images extracted from videos. Videos were recorded at night in clear weather, rain,
and in the fog, at different ranges, and with different movement types. YOLOv3 was significantly faster than
other detectors while achieving performance comparable with the best \cite{DBLP:journals/access/KristoIP20}.
In this paper, we mainly attacked the YOLOv3 model. After that, we transfered the infrared patch attack to other object detectors.

\section{Generating Infrared Adversarial Patches}

% In this section, we first give an overview of the adversarial attack problem and our proposed infrared 
% detection attack approach. We then introduce the loss function of the optimization model and the formula we 
% use. Next, we introduce the two patch models we built in the digital world, as well as the design and 
% manufacturing of the patch in the physical world.
We first formulate the problem, and introduce the attack method in the digital world, and then the attack 
method in the physical world.

\subsection{Problem Formulation}

We assume that the original image input is $x$, and the adversarial perturbation is $\delta $. 
Since we use a patch attack method, the perturbation only occupies part of the image. 
We assume the patched image is $\tilde{x}$.
% Since we use a patch attack method, the perturbation does not act on the whole image but occupies part of the image. We assume 
% that the proportion the perturbation occupied of the image is $\alpha \in \left( \text{0} , \text{1} \right)$, so 
% the image after adding perturbation can be described as
% \begin{equation}
%     \tilde{x}=\left( 1-\alpha  \right)\odot x+\alpha \odot \delta
% \end{equation}

% Suppose that the parameter of model $f$ is $\theta $, \(f\left( \theta ,\tilde{x} \right)\) represents the 
% output of the model with the input of perturbed image $\tilde{x}$. For the YOLO model, the output includes the 
% position of the bounding box \({{f}_{pos}}\left( \theta ,\tilde{x} \right)\), the object 
% probability \({{f}_{obj}}\left( \theta ,\tilde{x} \right)\), and the class 
% score \({{f}_{cls}}\left( \theta ,\tilde{x} \right)\). 
Let $f$ denote a model, $\theta $ denote its parameters, and \(f\left( x, \theta  \right)\) denote its output given the input $x$. 
Note that most object detector models have three outputs: position of of the bounding box \({{f}_{pos}}\left( x, \theta \right)\), the object probability \({{f}_{obj}}\left( x,\theta \right)\), and the class 
score \({{f}_{cls}}\left( x,\theta \right)\).  
Our goal is to attack the detection model 
so that the detection model can not detect objects of the specified category. In other words, we want to lower the \({{f}_{obj}}\left( x, \theta \right)\) score as much as possible. Therefore, the goal 
can be described as
\begin{equation}
    \min\limits_{\delta }\,{{f}_{obj}}\left( \tilde{x}, \theta  \right).
\end{equation}

% Our goal is to attack the object detector in the physical world, so we introduce the Expectation Over 
% Transformation(EOT) algorithm\cite{DBLP:conf/icml/AthalyeEIK18}. We need to consider various image transformations during the attack, 
% such as rotation, scale, noise, brightness, and contrast. The EOT algorithm makes the simulation process 
% more in line with physical reality. Assuming that the set of transformations is $T$, the goal can be 
% further described as 
% \begin{equation}
%     {{\tilde{x}}_{t}}=\left( 1-\alpha  \right)\odot x+\alpha \odot t(\delta )
% \end{equation}
% \begin{equation}
%     \min\limits_{\delta }\,{{\mathbb{E} }_{t\in T}}{{f}_{obj}}\left( \theta ,{{{\tilde{x}}}_{t}} \right)
% \end{equation}

% Furthermore, due to lots of intra-class variety of pedestrians, we hope to achieve a universal attack with multiple samples rather than one sample. Assuming that the data set has $N$ samples, our goal can be described as follows:
Our goal is to attack the object detector in the physical world, so we need to consider various image 
transformations of the patches during the attack, such as rotation, scale, noise, brightness, and contrast. Furthermore, 
due to lots of intra-class variety of pedestrians, we hope to achieve a universal attack on different people. 
Assuming that the set of transformations is $T$ , the patched image considering patch transformations 
is $ {{\tilde{x}}_{t}}$, and the data set has $N$ pedestrians, the goal can be described as
\begin{equation}
    \min\limits_{\delta }\,\frac{1}{N}\sum\nolimits_{i=1}^{N}{{{\mathbb{E} }_{t\in T}}{{f}_{obj}^{\left( i \right)}}\left( {{{\tilde{x}}}_{t}}, \theta \right)}.
    % \label{Lobj}
\end{equation}

% \subsection{Adversarial infrared attack loss}
% \noindent The goal of this work is to create a system that can fool infrared pedestrian detectors.Using an 
% optimization process, we try to find a patch that effectively lowers the accuracy of infrared person 
% detection on a large dataset. 
Our loss function consists of two parts:
\begin{itemize}
    \item ${{L}_{obj}}$ represents the maximum objectness score as shown in Equation (2).
    % The optimization goal is to make the score 
    % as low as possible. The lower the score, the lower the confidence that a specific target exists in the 
    % image. The goal of this work is to hide people by fooling the infrared pedestrian recognition system. 
    % \({{p}_{obj1}},{{p}_{obj2}},...,{{p}_{objn}}\) represent the objectness scores of the detection output.
    % We can calculate ${{L}_{obj}}$ as follows:
\end{itemize}
% \begin{equation}
%     {{L}_{obj}}=\max \left( {{p}_{obj1}},{{p}_{obj2}},{{p}_{obj3}},...,{{p}_{objn}} \right)
% \end{equation}

\begin{itemize}
    \item \({{L}_{tv}}\) represents the total variation of the image. This loss ensures that the optimized image 
    is smoother and prevents noisy images. We assume that ${{p}_{i,j}}$ represents the pixel value at a coordinate
        $\left( i,j \right)$ in the patch. We can calculate ${{L}_{tv}}$ as follows:
\end{itemize}
\begin{equation}
    {{L}_{tv}}=\sum\limits_{i,j}{\sqrt{{{\left( {{p}_{i,j}}-{{p}_{i+1,j}} \right)}^{2}}+{{\left( {{p}_{i,j}}-{{p}_{i,j+1}} \right)}^{2}}}}.
\end{equation}

We take the sum of the two losses weighted by the factor $\lambda $ which are determined empirically. Given by:
% \begin{equation}
%     L={{L}_{obj}}+\lambda {{L}_{tv}}
% \end{equation}

% % \subsection{Expectation over transformation (EOT)}
% The EOT algorithm proposed by Athalye et al. can effectively improve the robustness of the adversarial object. EOT is an approximation to the transformation in reality. It simulates a process in which an image is first printed by a printer and then taken by a camera. We consider a distribution $T$ that includes different transformations such as noise, contrast, rotation, and translation. We want a universal attack, which means that we can attack different people with the same patch. Pedestrians are different from objects with a fixed pattern, such as road signs. We assume that a pedestrian set contains N samples. The loss considering EOT is as follows:
\begin{equation}
\begin{split}
    L&={{L}_{obj}}+\lambda {{L}_{tv}}.\\
    % &=\max \left( {{p}_{obj1}},{{p}_{obj2}},{{p}_{obj3}},...,{{p}_{objn}} \right)+\lambda {{L}_{tv}}\\
%    &=\frac{1}{N}\sum\nolimits_{i=1}^{N}{{{\mathbb{E} }_{t\in T}}{{f}_{obj}^{\left( i \right)}}\left( \theta ,{{{\tilde{x}}}_{t}} \right)}\\
%    &+\lambda\sum\limits_{i,j}{\sqrt{{{\left( {{p}_{i,j}}-{{p}_{i+1,j}} \right)}^{2}}+{{\left( {{p}_{i,j}}-{{p}_{i,j+1}} \right)}^{2}}}}
\end{split}
\end{equation}

For the ensemble attack, We hope to lower the maximum objectness score of each detector at the same time. Assume 
there are $M$ detectors, and the maximum objectness score of $i$-th detector is $L_{obj}^{(i)}$ . We take the sum of these losses. So
the total loss of the ensemble attack is:
\begin{equation}
    {{L}_{ensemble}}=\sum\limits_{i=1}^{M}{L_{obj}^{(i)}}+ \lambda{{L}_{tv}}.
\end{equation}
% The overview of the proposed method is illustrated in Figure \ref{main_process}.
We use backpropagation to update the the patch iteratively (Figure \ref{main_process}).

\subsection{Digital Patch Attack}
% We built two kinds of digital patches, one is the pixel-level patch, the other is the Gaussian functions patch. 
% The digital patches are the simulations for the physical boardes.

\subsubsection{The Pixel-Level Patch}  
% The pixel-level patch uses pixels as the basic unit. Each pixel value is an 
% optimization variable. We follow the work of Simen et al.(2019) to build a square patch with the pixel size of
% $300\times300$. The difference is that our patch is a grayscale image instead of an RGB image. 
% The example is shown in Figure \ref{patch_2}(a). 
% We have considered using LED arrays to make pixel-level physical boardes. However, when we photographed the LED light array with an infrared thermal imaging camera, we found that 
% due to the low thermal efficiency of the LED light, it is difficult for the LED light array to show the 
% pattern we need in thermal imaging. 
First of all, we wondered if we follow the adversarial attack method on visible light images to fool pedestrian 
detectors\nocite{DBLP:conf/cvpr/ThysRG19}, what patch will be resulted in. Specifically, can the resulted patches be realized easily by 
using some thermal materials? It is easy to carry out this experiment because we only need to change the RGB 
images to grayscale images and follow the method described in \cite{DBLP:conf/cvpr/ThysRG19}. 
% The example is shown in Figure \ref{patch_2}(a). 

% Since the pixel value is positively correlated with the bulb temperature, the pixel value is also approximately to a Gaussian distribution.

% \begin{figure}[tb]
% \centering
% \includegraphics[width=1\columnwidth]{single_bulb_new2.jpg} % Reduce the figure size so that it is slightly narrower than the column. Don't use precise values for figure width.This setup will avoid overfull boxes. 
% \caption{The measured temperatures along the horizontal section shown in (a) are plotted in (b). The temperatures along other sections were nearly the same as those along the horizontal section, therefore are not shown in (b). }
% \label{single_bulb}
% \end{figure}

% \begin{figure}[htb]
% %\centering
% \begin{minipage}{90pt}
% %\centering
% \includegraphics[scale=0.39]{FLIR_large_65801_1X} % Reduce the figure size so that it is slightly narrower than the column. Don't use precise values for figure width.This setup will avoid overfull boxes. 
% \caption{the pixel-level patch}
% \label{fig2}   
% \end{minipage}
% \hspace{25pt}
% \begin{minipage}{90pt}
% %\centering
% \includegraphics[scale=0.39]{FLIR_large_N64_45628}
% \caption{the Gaussian functions patch}
% \label{fig3}
% \end{minipage}
% \end{figure}
% \begin{figure}[htb]
% \centering
% \includegraphics[width=0.94\columnwidth]{patch_compare.jpg} % Reduce the figure size so that it is slightly narrower than the column. Don't use precise values for figure width.This setup will avoid overfull boxes. 
% \caption{The digital patch}
% \label{patch_2}
% \end{figure}

\subsubsection{The Gaussian Functions Patch}  
To implement the adversarial attack in the physical world, another idea is to design an adversarial example 
based on the thermal property of certain given electronic components (e.g., resistors). We can first measure 
the relationship between the thermal properties of the components and the image patterns captured by the 
infrared cameras, then design an adversarial patch in the digital world, and finally manufacture a physical board specified by this digital adversarial patch.  Since infrared thermal imaging mainly uses the thermal radiation of the object, 
in the selection of electronic components, we consider diodes, resistors and small bulbs. 
% Due to the low thermal efficiency of diodes, we will not use diodes. Although the resistors are high in thermal efficiency, it is challenging to adjust the temperature intuitively. Therefore, we choose small bulbs. 
We found that the small bulb is a good candidate for adjusting image patterns captured by infrared cameras. Its brightness well relects its temperature. With the 
help of the rheostat, we can adjust the bulb brightness intuitively. We can fine-tune its infrared imaging pattern in this way.
% has not only better heating efficiency, but also its brightness can reflect its temperature.
% so that we can adjust the bulb’s brightness to change the thermal imaging effect. 
% We took an infrared image of a single bulb and found that the 
% temperature of the single bulb can be approximated to a  Gaussian distribution, as shown in Figure \ref{single_bulb}. 
% Therefore, we considered modeling the imaging of small bulbs and proposed the Gaussian functions patch.
\begin{figure}[tb]
\centering
\includegraphics[width=1\columnwidth]{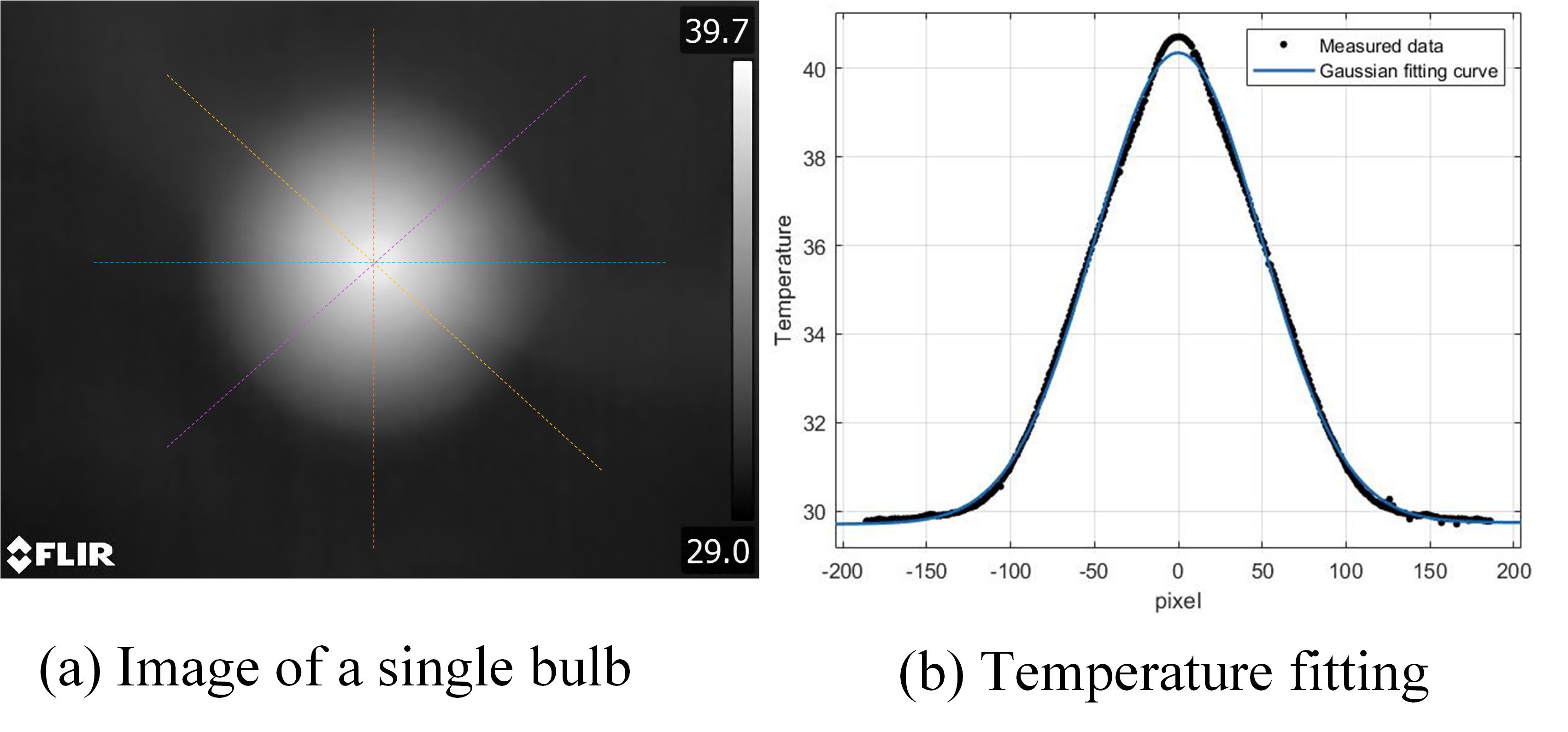} % Reduce the figure size so that it is slightly narrower than the column. Don't use precise values for figure width.This setup will avoid overfull boxes. 
\caption{The measured temperatures along the horizontal section shown in (a) are plotted in (b). The temperatures along other sections were nearly the same as those along the horizontal section, therefore are not shown in (b). }
\label{single_bulb}
\end{figure}

We took an infrared image of a single bulb. Then we used FLIR tools software provided by FLIR company to export the temperature of each point in 
the image. We first selected the temperature values on multiple lines that cross the center and then we used the Gaussian function for fitting as shown in Figure \ref{single_bulb}.
% After measurement, it is found that the temperature distribution of the bulb is isotropic.   
% Figure \ref{single_bulb} shows the measured data and Gaussian fitting curve. 
The fitting was good and the Root Mean Squard Error (RMSE) was 0.1511.
The amplitude value of the Gaussian function is 10.62, and the standard deviation is 70.07. %So we found that the Gaussian function can approximate the temperature of the single bulb.
Further experiments showed that the temperature of the same point measured by the infrared camera did not 
change with distance because of the correction function of the infrared camera. Therefore, the pixel value of 
the same point did not change with distance. 

% The pattern of the Gaussian functions patch is regarded as the superposition 
% of several Gaussian distributions. The optimized parameters for each Gaussian distribution include the position of 
% the Gaussian distribution center, the amplitude value of the Gaussian distribution, and the standard deviation. 
% In practice, the electrical parameters of the bulb are fixed. In order to meet the actual physical conditions, 
% the amplitude value and the standard deviation are obtained based on actual physical measurements.
% The advantage of this patch is in line with the radiation distribution of the point heat source in nature, so it 
% is more convenient for subsequent physical manufacturing. Besides, this kind of patch significantly reduces the 
% number of parameters compared with the pixel-level patch. In our experiment, the optimization parameters dropped 
% nearly 1000 times.
% The Gaussian functions patch is the foundation for subsequent physical manufacturing. 

% The pattern of the Gaussian functions patch is regarded as the superposition of several two-dimensional Gaussian functions. 
% In practice, the electrical parameters of the bulb are fixed. In order to meet the actual physical conditions, 
% the amplitude value and the standard deviation are obtained based on actual physical measurements. In our 
% experiment, the amplitude value is 0.863, and the standard deviation is 20. 
If we put many bulbs on a cardboard, the infrared camera will capture an image patch with a set of 2D Gaussian 
patterns, the centers of which have the highest pixel values.  The problem now is whether we can design such an 
image patch to fool the pedestrian detectors. We first carry out the attack using the Gaussian functions patch in 
the digital world. Since the amplitude and standard deviation of the Gaussian function is fixed to be measured values, the optimization paramter
of each two-dimensional Gaussian function is the coordinate of the center point. Besides, this kind of patch significantly reduces the number 
of parameters compared with the pixel-level patch. In our experiment, the number of optimization parameters dropped nearly 
1000 times.

Assuming that the pattern of a patch is superimposed by $M$ spots that conform to Gaussian functions, 
where the center point of the $i$-th Gaussian function is  $\left( {{p}_{x}},{{p}_{y}} \right)$, the 
amplitude amplification factor is ${{s}_{i}}$, and the standard deviation is ${{\sigma }_{i}}$. 
The measured ${{s}_{i}}$ was 10.62, and ${{\sigma }_{i}}$ was 70.07 in our experiment. 
We assume that the height of the entire image is $h$ , the width is $w$, and the coordinate of a single-pixel 
is $\left( x,y \right)$, where $x\in \left[ 0,w \right],y\in \left[ 0,h \right]$, then the $i$-th Gaussian 
function is as follows:
\begin{equation}
    {{g}^{(i)}}\left( x,y \right)={{s}_{i}}\cdot \exp \left( -\frac{{{\left( x-{{p}_{x}^{\left( i \right)}} \right)}^{2}}+{{\left( y-{{p}_{y}^{\left( i \right)}} \right)}^{2}}}{2\sigma _{i}^{2}} \right).
\end{equation}

Suppose the background of the patch is ${{P}_{back}}$, which is a 
matrix with all elements equal to $\mu $. The overall function of  multiple 2D Gaussian functions superimposed together is denoted by ${{P}_{syn}}$:
\begin{equation}
    {{P}_{syn}}={{P}_{back}}+\sum\limits_{i=1}^{M}{{{G}_{i}}}
\end{equation}

% \begin{equation}
%    {{P}_{back}}=\left( \begin{matrix}
%        \mu  & \ldots  & \mu   \\
%        \vdots  & \ddots  & \vdots   \\
%        \mu  & \cdots  & \mu   \\
%        \end{matrix} \right)     
% \end{equation}

\begin{equation}
    {{G}_{i}}=\left( \begin{matrix}
        {{g}^{(i)}}\left( 0,0 \right) & \ldots  & {{g}^{(i)}}\left( 0,w \right)  \\
        \vdots  & \ddots  & \vdots   \\
        {{g}^{(i)}}\left( h,0 \right) & \cdots  & {{g}^{(i)}}\left( h,w \right)  \\
        \end{matrix} \right).     
\end{equation}

\subsection{Physical Board Attack}
\begin{figure}[tb]
\centering
\includegraphics[width=0.95\columnwidth]{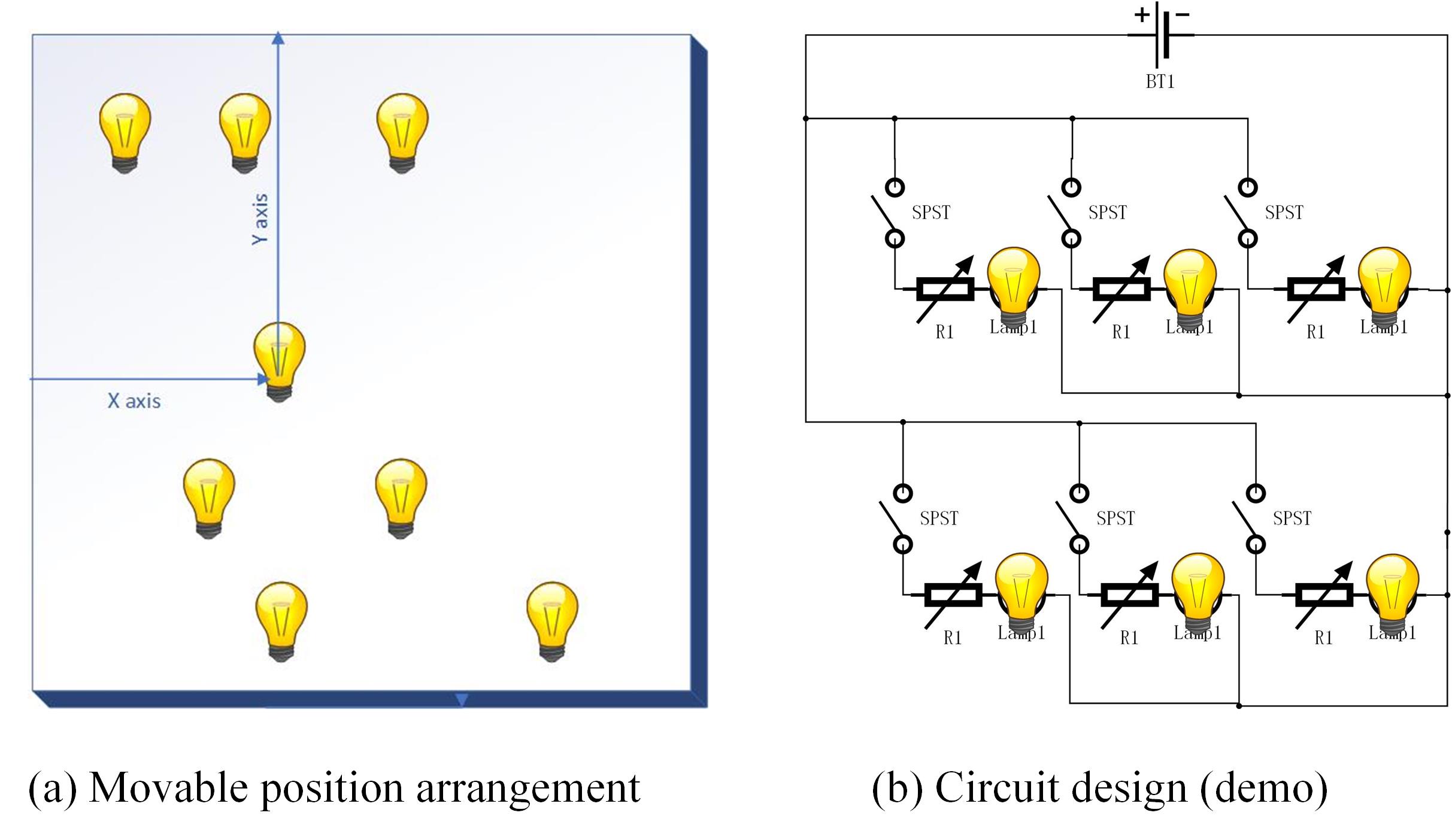} % Reduce the figure size so that it is slightly narrower than the column. Don't use precise values for figure width.This setup will avoid overfull boxes. 
\caption{Physical board design}
\label{patch_designs}
\end{figure}

In practice, we face a challenge to move the bulbs freely on a board when we try different patterns.
% changing the pattern, and to be fixed if the pattern needed to be unchanged. 
In other words, we need an adjustable physical board as shown in Figure \ref{patch_designs}(a).
We solve the problem by using magnets. One magnet is fixed to the 
bulb, and the other magnet is placed on the other side of the board. The magnet can attract the bulb on 
the other side and can adjust the position of the bulb as a button.

% In the design of the bulbs arrangement pattern, we have two designs: fixed position arrangement and movable arrangement. As shown in Figure \ref{patch_designs}(a) and Figure \ref{patch_designs}(b) respectively

% \begin{figure}[htbp]
% %\centering
% \begin{minipage}{90pt}
% %\centering
% \includegraphics[scale=0.32]{patch_design (5)} % Reduce the figure size so that it is slightly narrower than the column. Don't use precise values for figure width.This setup will avoid overfull boxes. 
% \caption{fixed position arrangement}
% \label{fig4}   
% \end{minipage}
% \hspace{30pt}
% \begin{minipage}{90pt}
% %\centering
% \includegraphics[scale=0.32]{patch_design (4)}
% \caption{movable arrangement}
% \label{fig5}
% \end{minipage}
% \end{figure}

% Simulation experiments show that the patch attack effect of fixed position arrangement is not as good as that of 
% the movable arrangement patch because the moveable patch could adjust not only the brightness and radiation 
% range of the bulb, but also the center of the light source. 

For circuit design, we used multiple independent power supply DC4.5V power modules. The rated voltage of the 
small bulb was 3.8V. After measurement, the total power of the physical board did not exceed 22W. 
The power supply lines of each light bulb were connected in parallel, and each circuit contained a small switch 
and a rheostat. This can ensure that the different small bulbs were independent of each other. 
The demo circuit design diagram is shown in Figure \ref{patch_designs}(b).

\section{Experiments and Results}
\subsection{Preparing the Data}
The dataset we used is FLIR\_ADAS dataset v1\_3 released by FLIR. FLIR\_ADAS provides an annotated thermal image
and non-annotated RGB image set for training and validation of object detection networks. The data set 
contains 10228 thermal images sampled from short videos and a continuous 144-second video. These photos were 
taken on the streets and highways in Santa Barbara, California, the USA from November to May. The thermal 
image is a FLIR Tau2 (13 mm f/1.0, 45-degree HFOV and 37-degree VFOV, FPA$640\times512$, NETD\textless60mK).

% \begin{figure*}[htbp]
% \centering
% \includegraphics[scale=0.39]{physical_examples_new.jpg} % Reduce the figure size so that it is slightly narrower than the column. Don't use precise values for figure width.This setup will avoid overfull boxes. 
% \caption{Gaussian functions patch attack and control experiments}
% \label{fig11}
% \end{figure*}

Thermal images are manually annotated and contain a total of four types of objects, namely \emph{people, bicycles, 
cars, and dogs}. Since we only care about \emph{people}, we filtered the original data set, and only kept images that contained people whose height is greater than 120 pixels. We finally selected 1011 images 
containing infrared pedestrians. We used 710 of them as the training set and 301 as the test set. We named 
them \textit{FLIR\_person\_select}.

\subsection{Target Detector}
% Mate et al.(2020) investigated the task of automatic person detection in thermal images
% using convolutional neural network models originally intended for detection in RGB images. They compared the
% performance of the standard state-of-the-art object detectors such as Faster R-CNN, SSD, Cascade R-CNN,
% and YOLOv3, that were retrained on a dataset of thermal images extracted from videos. Videos were recorded at night in clear weather, rain,
% and in the fog, at different ranges, and with different movement types. YOLOv3 was significantly faster than
% other detectors while achieving performance comparable with the best\cite{DBLP:journals/access/KristoIP20}.
Mate et al.(2020) have compared the performance of the standard state-of-the-art infrared object detectors such as Faster R-CNN, SSD, Cascade R-CNN,
and YOLOv3. They found that YOLOv3 was significantly faster than other detectors while achieving performance comparable with the best.
So we chose YOLOv3 as the target detector. The network has 105 layers. We resized the input images to $416\times416$ as required by the model. We chose the pre-training weights officially provided by YOLO and then fituned on 
\textit{FLIR\_person\_select}. The AP of the model was 0.9902 on the training 
set, and 0.8522 on the test set. We used this model as the target model of attack.

% \begin{figure}[htb]
% \centering
% \includegraphics[width=0.8\columnwidth]{pr-curve-large_FLIR_65801_1X_hb0.75} % Reduce the figure size so that it is slightly narrower than the column. Don't use precise values for figure width.This setup will avoid overfull boxes. 
% \caption{evaluation of pixel-level patch attack}
% \label{fig12}
% \end{figure}

% \begin{figure}[htb]
% \centering
% \includegraphics[width=0.8\columnwidth]{pr-curve-large_FLIR_N64_45628_1X_hb0.75} % Reduce the figure size so that it is slightly narrower than the column. Don't use precise values for figure width.This setup will avoid overfull boxes. 
% \caption{evaluation of Gaussian functions patch attack}
% \label{fig13}
% \end{figure}

% \subsection{Evaluation of attacks in the digital world}
\subsection{Simulation of Physical Attacks}
% In the section \textit{The digital patch}, we mentioned that we designed two kinds of digital patches. One uses 
% pixels as the basic unit, and the pixel value of each pixel is an optimization variable. The other uses the spot 
% in line with gauss distribution as the basic unit, and the parameters of each Gaussian model are optimization 
% variables. In this section we will study the attack effects of these two patches on pedestrian detection. 
% These experiments are the simulations of physical attacks. The workflow of training patch and physical board 
% attack is shown in Figure \ref{fig18}. We first use pixel-level patches to verify the feasibility of infrared 
% image attacks, and then use Gaussian functions patches to simulate physical attacks. Finally, the physical board 
% pattern is determined according to the simulation results.

\subsubsection{Pixel-Level Patch Attack}
Following the process described by Simen et al. (2019), we obtained a patch shown in Figure \ref{patch_2} (a).
The attack was successful as the patch made the accuracy of YOLOv3 dropped by 74.57\% (see Supplementary Information 
for more details). However, the resulted patch contained numourous grayscale pixels as noise, which are difficult 
to realize physically. Therefore we abandoned this approach. 
\begin{figure}[tb]
\centering
\includegraphics[width=1\columnwidth]{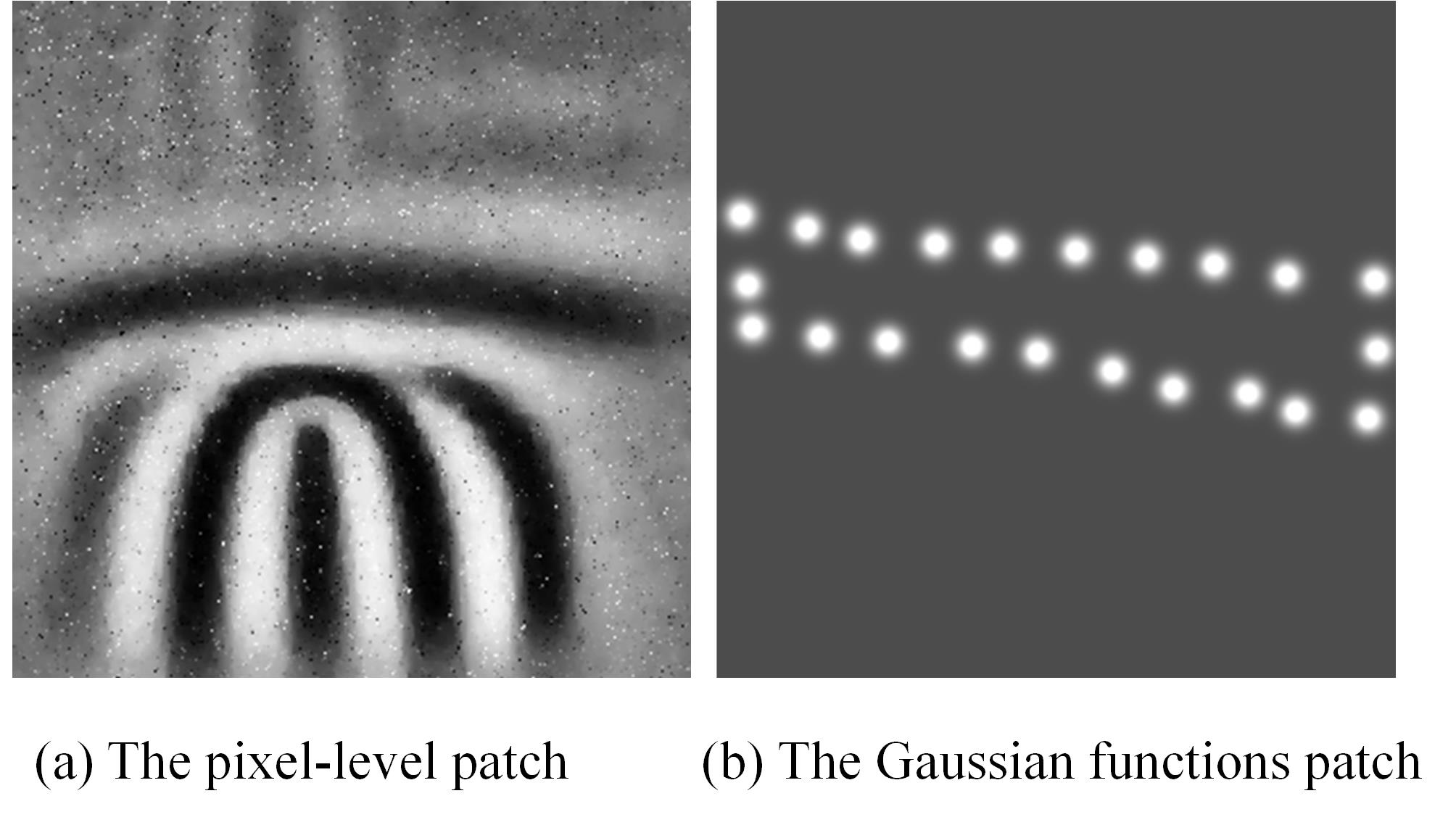} % Reduce the figure size so that it is slightly narrower than the column. Don't use precise values for figure width.This setup will avoid overfull boxes. 
\caption{The digital patch}
\label{patch_2}
\end{figure}
% Next, we apply the optimized patch to the \textit{FLIR\_person\_select} 
% test set. We also designed two patches as controls: One is a random noise patch with the same size, which we named 
% random\_patch; the other is a uniform gray patch, we name it hot\_board because it corresponds to the thermal 
% image of a board with a constant temperature in the real world.We adopt the IOU method to calculate the accuracy 
% of the detection. As for the results, we show the Precision-Recall curve in Figure \ref{fig12}. 
% We see that using the output of the clean image input as ground truth, the accuracy of network detection 
% is reduced by 25.30\% and 30.51\%, using random\ noise patch and using hot\_borad patch respectively, while the 
% pixel-level patch we designed makes the accuracy of network detection drop by 74.57\%. We give some examples 
% to see the effect of attacks intuitively in Figure \ref{fig10}. 
% world.
% patch with the same size, which we named random\_patch; the other is a uniform gray patch, we name it 
% hot\_board
% The Figure \ref{patch_2}(a) is the patch obtained after 66439 iterations.

% \begin{figure*}[htbp]
% \centering
% \includegraphics[scale=0.38]{FLIR_pixel_example} % Reduce the figure size so that it is slightly narrower than the column. Don't use precise values for figure width.This setup will avoid overfull boxes. 
% \caption{Pixel-level patch attack and control experiments}
% \label{fig10}
% \end{figure*}
\begin{figure}[bp]
\centering
\includegraphics[width=0.8\columnwidth]{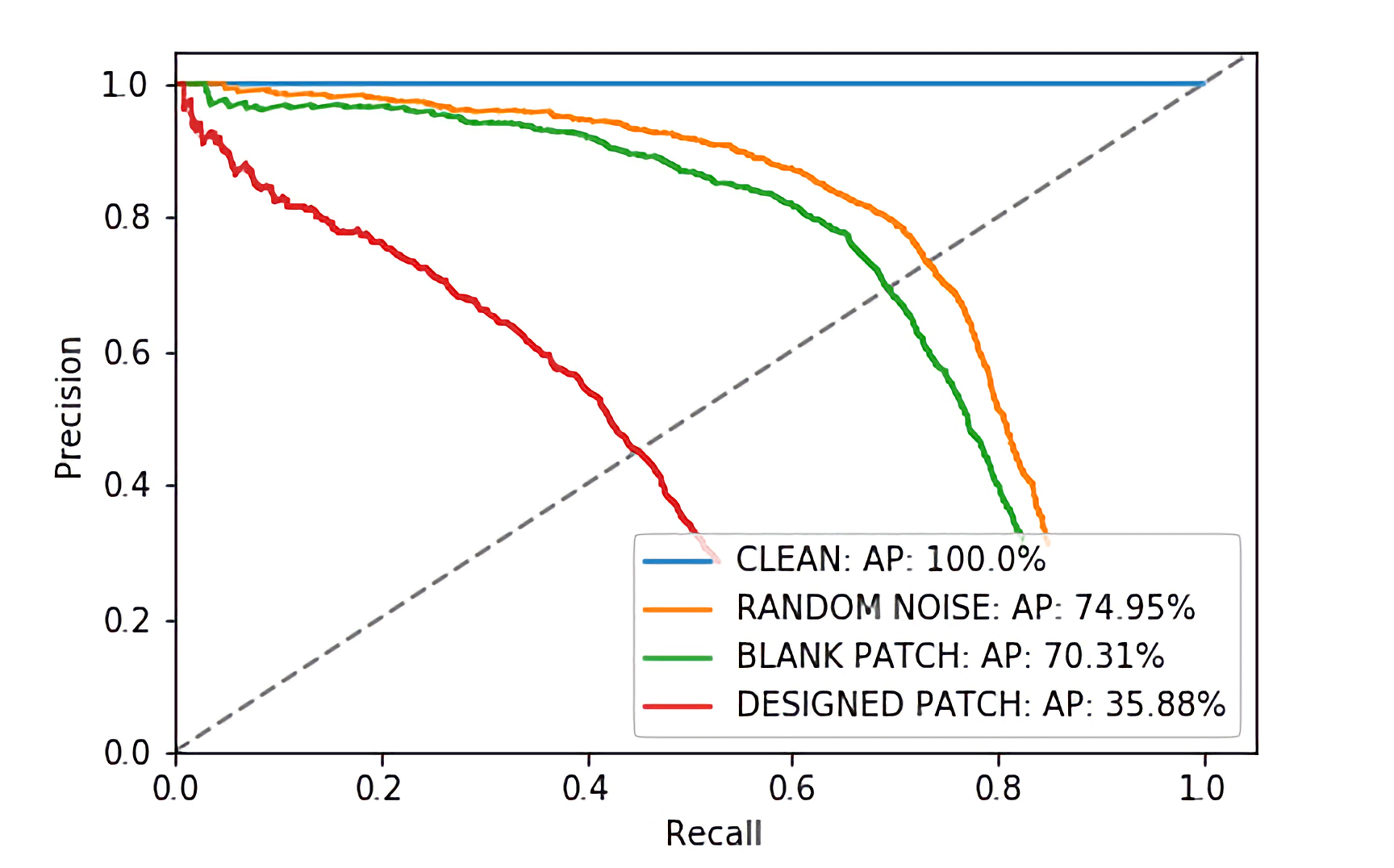} % Reduce the figure size so that it is slightly narrower than the column. Don't use precise values for figure width.This setup will avoid overfull boxes. 
\caption{Evaluation of Gaussian functions patch attack}
\label{fig13}
\end{figure}

\begin{figure*}[tbp]
\centering
\includegraphics[scale=0.42]{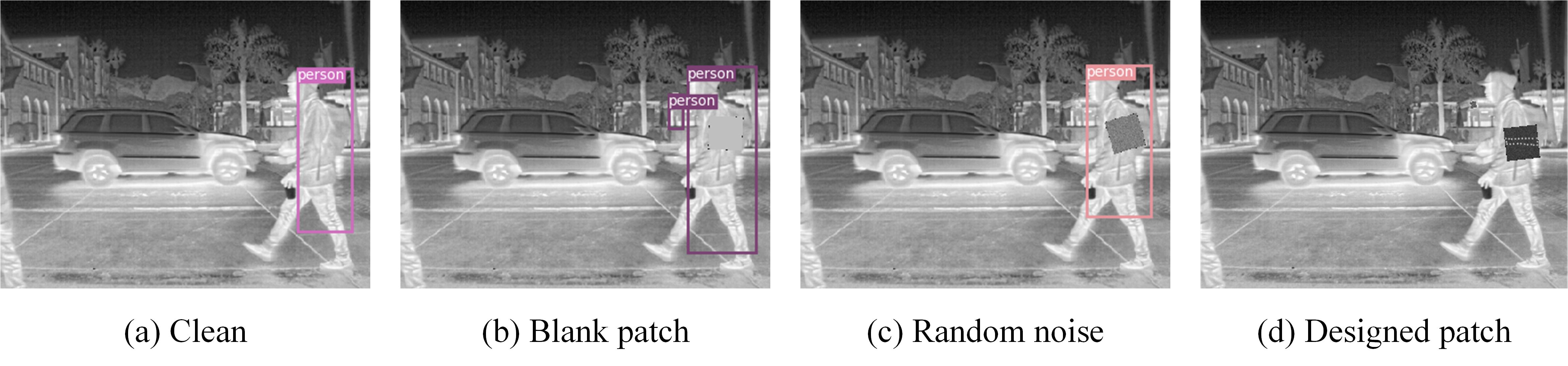} % Reduce the figure size so that it is slightly narrower than the column. Don't use precise values for figure width.This setup will avoid overfull boxes. 
\caption{Gaussian functions patch attack and control experiments}
\label{fig11}
\end{figure*}

\subsubsection{Gaussian Functions Patch Attack}
The pattern of the Gaussian functions patch is superimposed by multiple spots that conform to a two-dimensional 
Gaussian function (Figure \ref{single_bulb}). 
% Each spot can be determined by position coordinates, amplitude value, and standard deviation. 
% We have two optimization strategies. One is that the spot position is fixed, and only the amplitude 
% and standard deviation are optimized. The other is that the spot position can be changed, and all the parameters 
% of the Gaussian model could be optimized. In our experiments, we found that the attack effect of the second 
% strategy is better than the former, so we adopt a Gaussian model with the moveable spots.
% The method of generating the Gaussian functions patch is similar to the pixel-level patch. The difference is that the 
% pixel-level patch uses an optimizer to update the value of each pixel, while the Gaussian functions patch uses an 
% optimizer to update the parameters of each Gaussian function. When the patch size is $300\times300$, the 
% pixel-level patch needs to optimize 90,000 parameters, while the Gaussian functions patch with 30 spots needs to optimize 
% only 120 parameters. Therefore, the Gaussian functions patch dramatically reduces the number of optimized parameters 
% and also lays the foundation for subsequent physical implementation. 
% Note that we only need to optimize the locations of the Gaussian functions. If we assign 30 Gaussian functions 
% in the patch, there are only 60 parameters. This number is significantly smaller than the 
% number of parameters needed to be optimized in the pixel-level patch attack. 
To make the patch more robust, we designed a variety of transformations including  
random noise on the patch, random rotation of the patch (clockwise or counterclockwise within 20 degrees), 
random translation of the patch, and random changes in the brightness and contrast of the patch. These 
transformations simulate the perturbation of the physical world to a certain extent, which effectively 
improves the robustness of the patch. 
Then we used the training set of \textit{FLIR\_person\_select} and placed the patch on the upper body of the pedestrians according to the position of the bounding box. 
The size of the patch was 1/5 of the height of the bounding box.
Next, we used the patched image as input and ran the YOLOv3 pedestrian detector we had trained. 
% The detector will output the score of the pedestrian. Our goal is to make 
% it as low as possible. Then the optimizer uses the backpropagation algorithm to update the pixel value of 
% each pixel so that the new patch can better fool the detector. We show the process in Figure \ref{main_process}. 
We used a stochastic gradient descent optimizer with momentum, and the size of each batch was 8. The optimizer 
used the backpropagation algorithm to update the parameters of 
Gaussian functions by minimizing Equation (4).%so that the new patch can better fool the detector.
% Before training the patch, we assumed that the number of spots is between 9 and 81. During the optimization process, the number of spots and 
% the position of the spots were continually changing to reduce the loss function. 
 Through this process, we obtained a series of patches with different numbers of Gaussian functions. 
Figure \ref{patch_2}(b) is an example with 22 Gaussian functions. 
% We used Precision-Recall(P-R) curve like Figure \ref{fig13} to evaluate the effect of patch attack. 
% Average precision(AP) is a measure that combines recall and precision.
% We evaluated the attack effect of different patches by AP, the results is shown in Table \ref{tab1}.
% \begin{table}[!htbp]
% \centering
% \caption{The study of the number of Gaussian functions}\label{tab1}
% \begin{tabular}{cc}
% \toprule  
% The number & The AP dropped by \\
% \midrule  
% 9& 42.7\%\\
% 15& 56.0\%\\
% 22& 64.5\%\\
% 25& 64.9\%\\
% 36& 66.2\%\\
% \bottomrule 
% \end{tabular}
% \end{table}

% When the loss gradually converged, we found that the trained patches had a similar pattern like Figure \ref{patch_2}(b). 
Next, we applied the optimized patch which is shown in Figure \ref{patch_2}(b) to the test set, using the same process we used during 
training, including various transformations. 
% We also designed two patches as controls: One is a random noise
% patch with the same size, which we named random\_patch; the other is a uniform gray patch, we name it 
% hot\_board. 
We used random noise patches with maximum amplitude value 1 and constant pixel value patches (blank patches) for control experiments. 
The pixel values of blank patch in our experiment were 0.75. We tried other values and found that blank patches with different pixel values had a similar attack effect.
% The hot board patchis namely blank patch. It is the simulation of hot board with the uniform temperature in the physical world.
We applied these different patches to the \textit{FLIR\_person\_select} test set, and then input 
the patched images to the same detection network to test its detection performance. 
% One advantage of our method is that there is no need for manual annotation when testing because the bounding 
% boxes obtained from clean images input are the ground truth. 
We adopt the IOU method to calculate the accuracy 
of the detection. The precision-recall (PR) curves are shown in Figure \ref{fig13}. 
% Average precision is a measure that combines recall and precision.
% The general definition for the Average Precision (AP) is finding the area under the precision-recall curve above.
% As for the results, we show the Precision-Recall curve in Figure \ref{fig12}:
% We apply the Gaussian functions patch to \textit{FLIR\_person\_select} to
% evaluate the attack effect. We use random noise patches and constant gray value patches(hot\_board patches) for control experiments. 
% 
% \begin{figure}[htbp]
% \centering
% \includegraphics[width=0.7\columnwidth]{pr-curve-large_FLIR_N64_45628_1X_hb0.75} % Reduce the figure size so that it is slightly narrower than the column. Don't use precise values for figure width.This setup will avoid overfull boxes. 
% \caption{Evaluation of Gaussian functions patch attack}
% \label{fig13}
% \end{figure}
% 
% \begin{figure*}[htbp]
% \centering
% \includegraphics[scale=0.45]{physical_examples2.jpg} % Reduce the figure size so that it is slightly narrower than the column. Don't use precise values for figure width.This setup will avoid overfull boxes. 
% \caption{physical board attacks and control experiments}
% \label{fig17}
% \end{figure*}
% 
Using the output of the clean image input as ground truth, the Gaussian fuctions patch we designed made the 
average precision (AP, the area under the PR curve) of the target detector drop by 64.12\%. We give an example to show the attacking effect in Figure \ref{fig11}.
In contrast, the AP of the target detector dropped by 25.05\% and 29.69\% using random noise patch and 
using blank patch, respectively. 
% the accuracy of network detection was 
% reduced by 24.05\% and 30.30\%, using random noise patch and using hot\_borad patch respectively, while the Gaussian 
% model patch we designed makes the accuracy of network detection drop by 64.5\%. 

Note that the attack performance of the Gaussian functions patch attack was not as good as the pixel-level 
patch attack. This is reasonable as the latter had nearly 1000 times more parameters than the former. But 
the former is easier to be realized physically. 

We tried different kinds of patches and evaluated the attack effect of different patches by AP, the results is shown in Table \ref{tab1}.
\begin{table}[!tbp]
\centering
% \caption{The study of the number of Gaussian functions}\label{tab1}
\begin{tabular}{cc}
\toprule  
The number & The AP dropped by \\
\midrule  
9& 46.02\%\\
15& 51.26\%\\
22& 64.12\%\\
25& 65.74\%\\
36& 66.88\%\\
\bottomrule 
\end{tabular}
\caption{The study of the number of Gaussian functions}\label{tab1}
\end{table}
We found the Gaussian functions patch with 22 spots had a good attack effect, while maintaining a small amount of
parameters. 

\subsection{The Effects of Patch Size}
\begin{figure}[b]
\centering
\includegraphics[width=0.8\columnwidth]{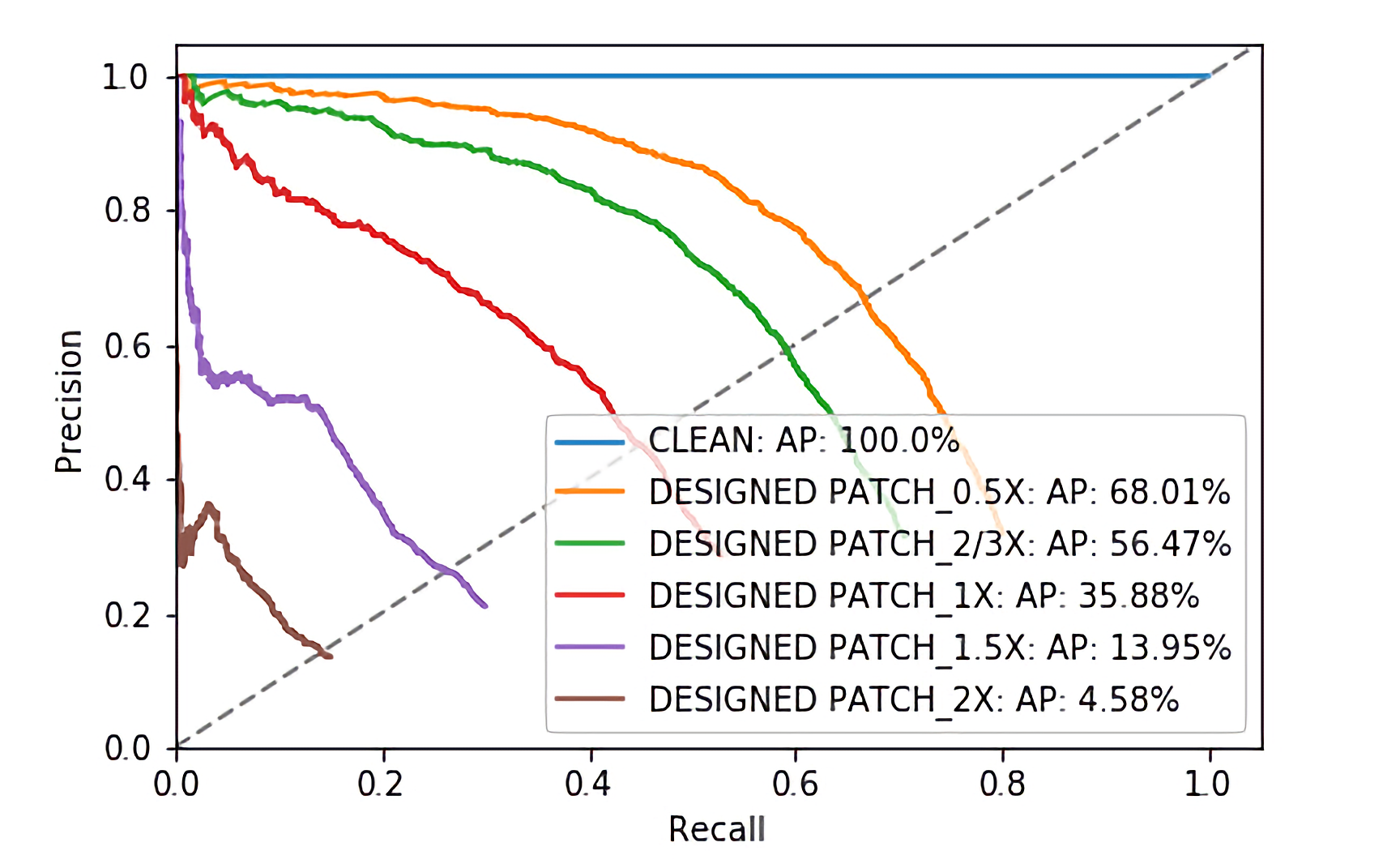} % Reduce the figure size so that it is slightly narrower than the column. Don't use precise values for figure width.This setup will avoid overfull boxes. 
\caption{Evaluation of patch attack with different size}
\label{fig19}
\end{figure}

We scaled up or down the original image to study the effect of patch 
size on the attack. We used the patch shown in Figure \ref{patch_2}(b). We did five experiments. One kept 
the original size of the patch ($300\times300$), the 
other two expanded the side length by 1.5 times and 2 times respectively, and the last two reduced the 
side length to 2/3 and 1/2 of the original respectively. The results are shown in Figure \ref{fig19}. The patch which is doubled in size caused AP of YOLOv3 to drop by 95.42\%.
We found that when the patch size dropped to 1/2 of its original size, its attack performance dropped a lot. 
That's the limit of our patch attack method.

\subsection{Evaluation of Attacks in the Real World}
\begin{figure}[!b]
\centering
\includegraphics[width=0.95\columnwidth]{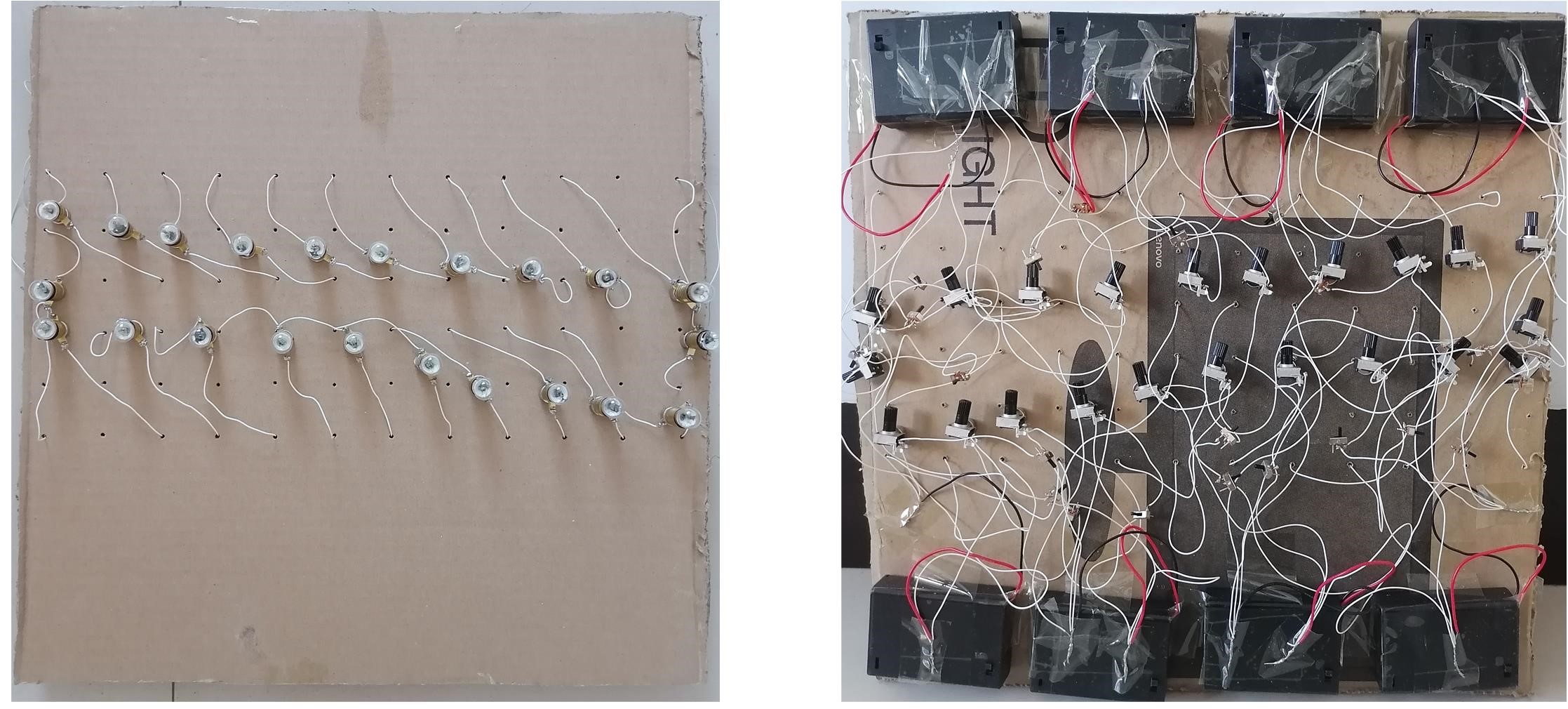} % Reduce the figure size so that it is slightly narrower than the column. Don't use precise values for figure width.This setup will avoid overfull boxes. 
\caption{Physical board we manufactured}
\label{physical_patch}
\end{figure}

\begin{figure}[!b]
\centering
\includegraphics[width=0.95\columnwidth]{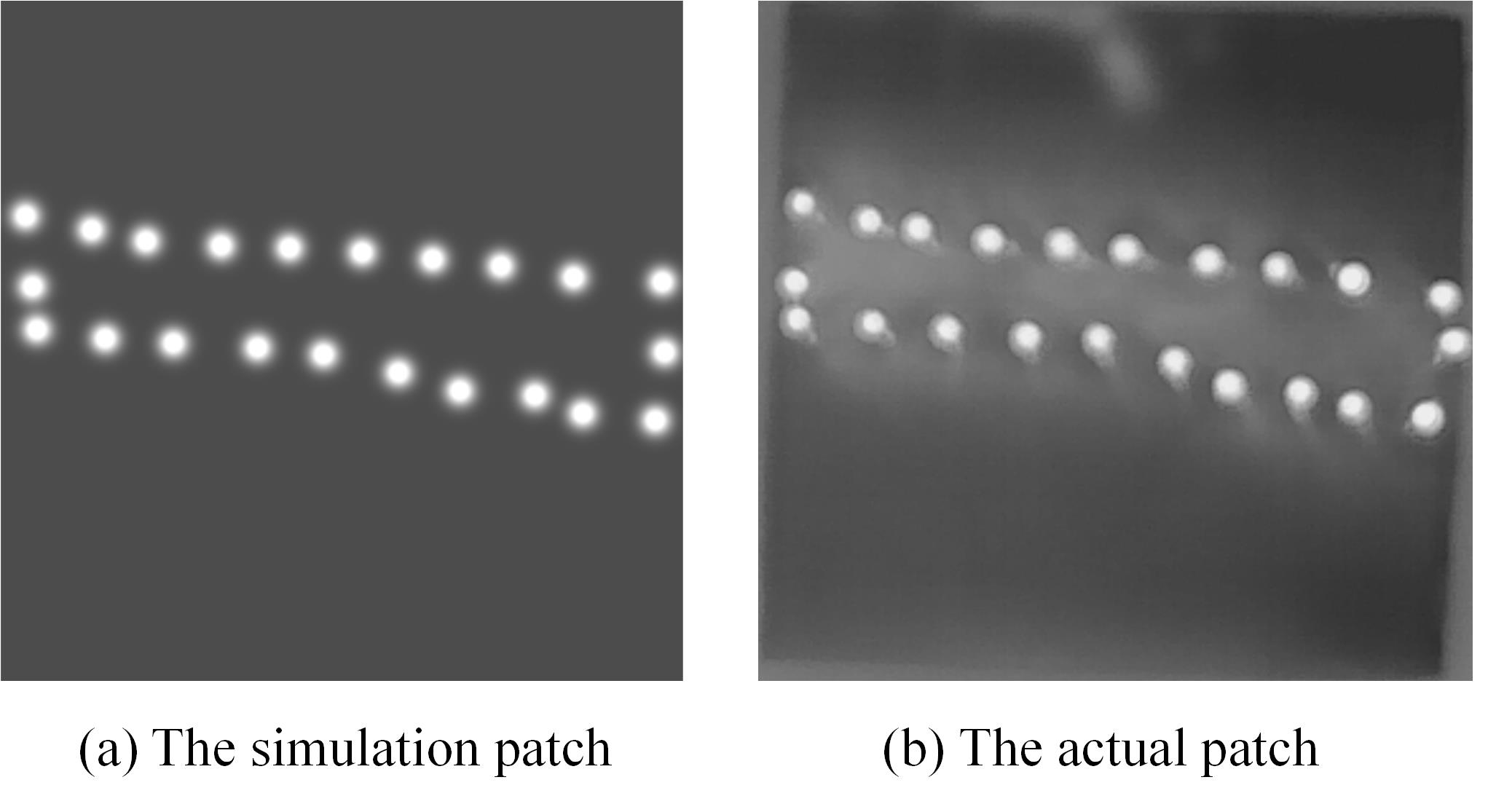} % Reduce the figure size so that it is slightly narrower than the column. Don't use precise values for figure width.This setup will avoid overfull boxes. 
\caption{Comparation between simulated and actual board}
\label{compare}
\end{figure}

% Our goal is to achieve physical attacks on pedestrian detectors in the real world. We introduced how we make 
% the physical board in detail in the section \textit{The physical board}. 
The pattern of the physical board is derived from the Gaussian functions patch as shown in Figure \ref{patch_2}(b). 
We chose a 35cm$\times$35cm cardboard. The finished board is shown in Figure \ref{physical_patch}. It is worth 
noting that the total manufacturing cost of our physical board did
not exceed \$5, indicating the proposed approach is economic. 
Figure \ref{compare} shows the simulated and actual boards.

% Before conducting physical attacks, we performed simulations in the digital 
% world. In order to make the simulation more in line with reality, we further optimize the design of the 
% Gaussian functions patch according to the actual thermal photos. We fine-tune the parameters of the Gaussian 
% model patch based on the patch shown in Figure \ref{patch_2}(b). The Figure \ref{compare}(a) and Figure \ref{compare}(b) show the comparison between the simulation 
% patch and the actual physical board.

% \begin{figure*}[htbp]
% \centering
% \includegraphics[scale=0.45]{physical_examples2.jpg} % Reduce the figure size so that it is slightly narrower than the column. Don't use precise values for figure width.This setup will avoid overfull boxes. 
% \caption{physical board attacks and control experiments}
% \label{fig17}
% \end{figure*}

\begin{figure*}[htbp]
\centering
\includegraphics[scale=0.39]{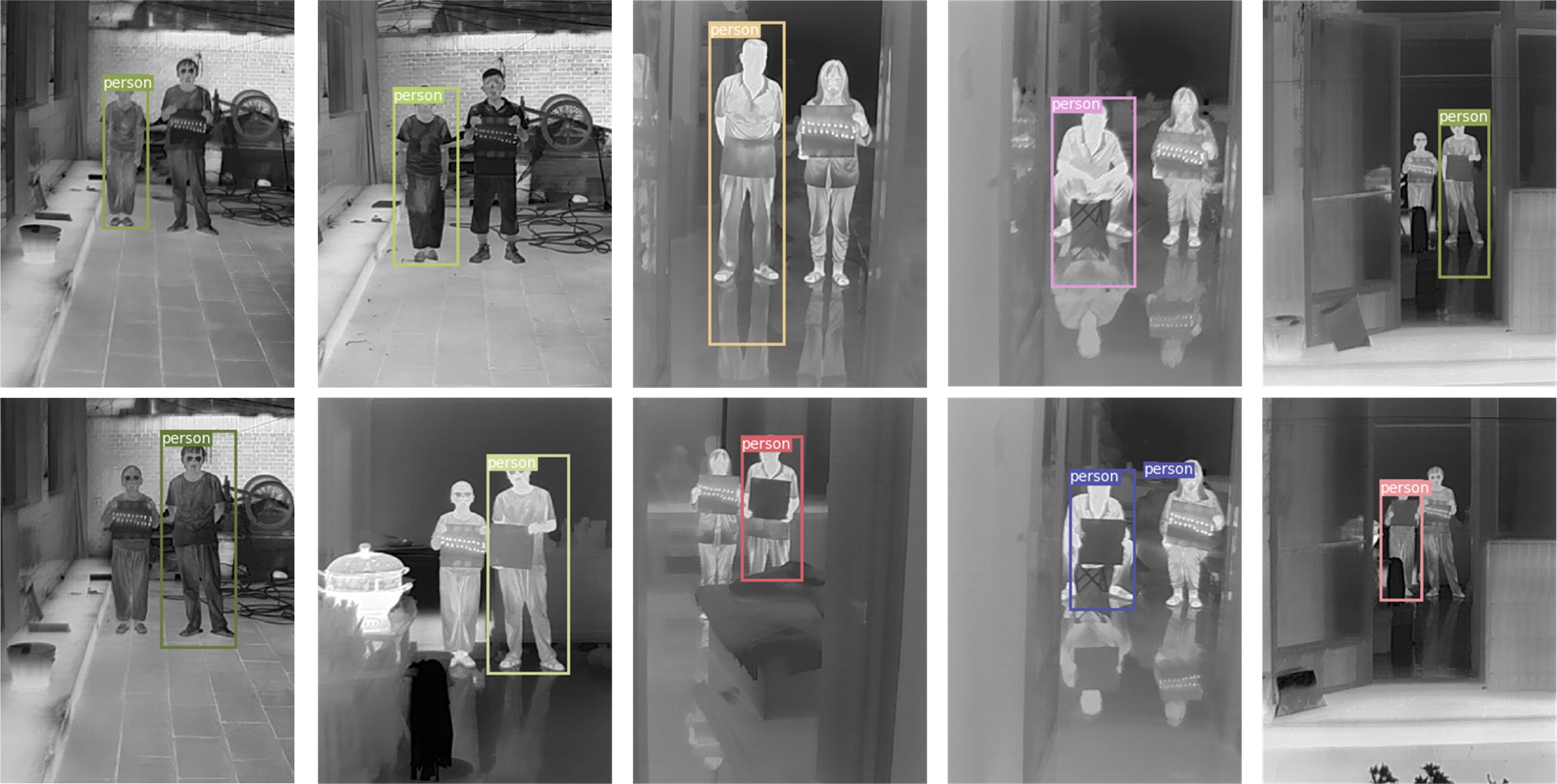} % Reduce the figure size so that it is slightly narrower than the column. Don't use precise values for figure width.This setup will avoid overfull boxes. 
\caption{Physical board attacks and control experiments}
\label{fig17}
\end{figure*}

We conducted physical attack experiments. The equipment we used was HTI-301 infrared camera (FPA $384\times288$, NETD\textless 60mK). 
We invited several people to do the experiment. They could hold the adversarial board, 
or a blank board, or nothing.  We used the infrared camera to shoot these people at the same time under the same conditions, and then sent the thermal 
infrared photos to the pedestrian detector for testing. 
Some examples of testing results are shown in Figure \ref{fig17}.
% 
% \begin{figure*}[htbp]
% \centering
% \includegraphics[scale=0.45]{physical_examples2.jpg} % Reduce the figure size so that it is slightly narrower than the column. Don't use precise values for figure width.This setup will avoid overfull boxes. 
% \caption{physical board attacks and control experiments}
% \label{fig17}
% \end{figure*}
% 
It is seen that whenever a person held the blank board or nothing, YOLOv3 could detect her/him. However, 
if a person held the adversarial board, YOLOv3 could not detect her/him. The results show 
the physical board we designed can successfully attack the infrared pedestrian detection system in the 
real world. 
% This is the first work that we know to achieve physical attacks on pedestrians in the field of 
% thermal infrared images.

To quantify the effects of a physical attack, we recorded 20 videos from different scenes(See supplementary materials for the demo video). 
We invited several people to be the actors of the videos. For fair comparison, we asked the actors to walk 
three times from the same starting position to the same end positions with the same path, 
once with the adversarial board, once with the blank board and another with nothing. Each group of videos were taken in the same condition. Ten videos were recorded 
indoors, and the others were recorded outdoors. Each video took 5-10 seconds. The camera got 30 frames per 
second. We considered different distances (between 3 and 15 meters)
and angles (from 45 degrees on the left to 45 degrees on the right) when recording the video. 
There were 1853 frames of images that contained the physical board we designed, 1853 frames of images that contained the blank board, and 
the other 1853 frames that just contained pedestrians. 
% We set the threshold of the pedestrian detector to 0.8. Without the physical board, 
% there were 1796 frames of images that had the identified pedestrian. Therefore, the pedestrian detection rate 
% for a clean image was 1796/1853 = 96.9\%. When the physical board was applied, pedestrians were detected in 973 
% frames of images. So the success rate of physical infrared patch attack can be calculated as (1796-973)/1796= 45.8\%.
The result showed that the cardboard caused the AP of the target detector to drop by 34.48\%, while a blank 
board with the same size caused the AP to drop by 14.91\% only.
\begin{figure}[!b]
\centering
\includegraphics[width=0.95\columnwidth]{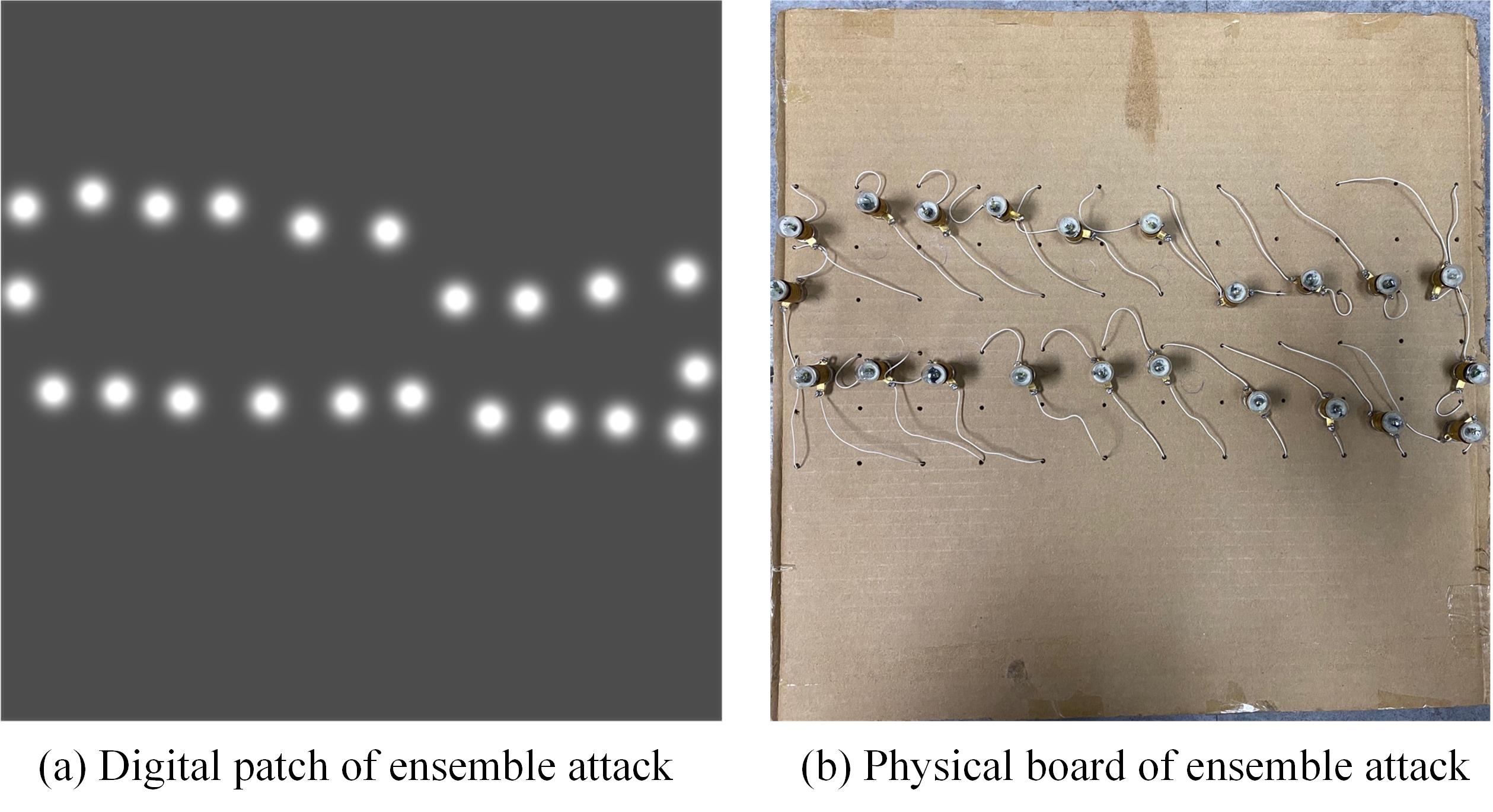} % Reduce the figure size so that it is slightly narrower than the column. Don't use precise values for figure width.This setup will avoid overfull boxes. 
\caption{The patch obtained by model ensemble}
\label{embed_patch}
\end{figure}
% \subsection{Further experiments}
% \subsection{The effects of patch size}
% With the pattern remaining unchanged, we scaled up or down the original image to study the effect of patch 
% size on the attack. We used the patch shown in Figure \ref{patch_2}(b). We did five comparative experiments. One keeps the original size of the patch, the 
% other two expands the side length by 1.5 times and 2 times respectively, and the last two reduced the 
% side length to 2/3 and 1/2 of the original respectively. The results are shown in Figure \ref{fig19}.

% \begin{figure}[htb]
% \centering
% \includegraphics[width=0.8\columnwidth]{pr-curve-FLIR_N64_scale.png} % Reduce the figure size so that it is slightly narrower than the column. Don't use precise values for figure width.This setup will avoid overfull boxes. 
% \caption{Evaluation of patch attack with different size}
% \label{fig19}
% \end{figure}

\subsection{Ensemble Attack}
In order to study whether we can transfer the infrared patch attack to other detection models, we did the following experiment. At the beginning, we directly 
used the patch which successfully attacked YOLOv3 to attack other detectors, such as Cascade-RCNN and RetinaNet. 
The patch trained on YOLOv3 caused the AP of Cascade-RCNN and RetinaNet to drop by 11.60\% and 25.86\%. To improve the transferability of attack, we use model ensemble techniques.
% We integrated YOLOv3, Faster-RCNN, and Mask-RCNN during training. Then we trained a new Gaussian patch. 
We obtained a new Gaussian patch by integrating YOLOv3, Faster-RCNN, and Mask-RCNN during training as shown in Figure \ref{embed_patch}. 
% Our targets are Cascade-RCNN and RetinaNet. 
% We first attacked in the digital world. The results are shown in Figure \ref{tra_dig}(a) and Figure \ref{tra_dig}(b) respectively.
% Our patch make the average precision of Cascade RCNN and RetinaNet dropped by 35.73\% and 46.95\% respectively in the digital world.
Cascade-RCNN and RetinaNet were attacked in the digital world firstly. As shown in Table \ref{tab2}, 
our patch caused the AP of Cascade RCNN and RetinaNet to drop by 35.28\% and 46.95\% respectively, 
% while the patch trained on YOLOv3 caused the AP to drop by 12.60\% and 25.86\% respectively.
% \begin{figure}[htb]
% \centering
% \includegraphics[width=1\columnwidth]{transfer_digital_attack.jpg} % Reduce the figure size so that it is slightly narrower than the column. Don't use precise values for figure width.This setup will avoid overfull boxes. 
% \caption{Transferability in the digital world}
% \label{tra_dig}
% \end{figure}
\begin{table}[!bp]
\centering
% \caption{Transferability in the digital world}\label{tab2}
\begin{tabular}{c|cc}
\toprule  
\diagbox{Train}{Test} & Cascade RCNN & RetinaNet \\
\midrule  
YOLOv3& 11.60\% &25.86\%\\
\midrule 
YOLOv3+Faster\\-RCNN+Mask-RCNN& 35.28\% &46.95\%\\
\bottomrule 
\end{tabular}
\caption{Transferability in the digital world}\label{tab2}
\end{table}
After that, we conducted experiments in the physical world, and the experimental settings were consistent with section \textit{Evaluation of attacks in the real world}. 
Our patch make the AP of Cascade RCNN and RetinaNet 
dropped by 21.67\% and 25.17\% respectively in real world.
% while the blank board with the same size caused the AP to drop by 11.41\% 
% and 12.56\% respectively.
% \begin{figure}[htb]
% \centering
% \includegraphics[width=1\columnwidth]{transfer_physical_attack.jpg} % Reduce the figure size so that it is slightly narrower than the column. Don't use precise values for figure width.This setup will avoid overfull boxes. 
% \caption{Transferability in the physical world}
% \label{tra_phy}
% \end{figure}
% \begin{table}[!htbp]
% \centering
% \caption{Transferability in the physical world}\label{tab3}
% \begin{tabular}{cc}
% \toprule  
% Target model(black-box) & The AP dropped by \\
% \midrule  
% Cascade RCNN& 21.60\%\\
% RetinaNet& 25.17\%\\
% \bottomrule 
% \end{tabular}
% \end{table}

\section{Conclusion}
In this article, we demonstrate that it is possible to attack the infrared pedestrian detector in the real world. 
We propose two kinds of patches: the pixel-level patch and the Gaussian functions patch. 
% We implemented the physical board attack using the physical patch we designed and manufactured. 
We implement the attack in the physical world by designing a cardboard decorated with small bulbs. 
% Our physical adjustable patch can switch different patterns at any time according to different scenes. 
The physical board can successfully fooled the infrared pedestrian detector based on YOLOv3
. In addition, by using the ensemble attack technique, we designed a cardboard that could fool detectors that were unknown to us. 
As the thermal infrared detection systems are widely used in night 
security, automatic driving, especially body temperature detection during COVID-19, our work has important 
practical significance. 

% \  \textbf{Acknowledgements.}\  
\section{Acknowledgements}
This work was supported in part by National Natural Science Foundation of China under Grant 61734004, 
Grant U19B2034, Grant 62061136001 and Grant 61620106010. %and in part by the NSFC/DFG Collaborative Research Centre SFB/TRR169 "Crossmodal Learning" II.
\bibliography{myref1}

\begin{thebibliography}{26}
\providecommand{\natexlab}[1]{#1}
\providecommand{\url}[1]{\texttt{#1}}
\providecommand{\urlprefix}{URL }
\expandafter\ifx\csname urlstyle\endcsname\relax
  \providecommand{\doi}[1]{doi:\discretionary{}{}{}#1}\else
  \providecommand{\doi}{doi:\discretionary{}{}{}\begingroup
  \urlstyle{rm}\Url}\fi

\bibitem[{Athalye et~al.(2018)Athalye, Engstrom, Ilyas, and
  Kwok}]{DBLP:conf/icml/AthalyeEIK18}
Athalye, A.; Engstrom, L.; Ilyas, A.; and Kwok, K. 2018.
\newblock Synthesizing Robust Adversarial Examples.
\newblock In Dy, J.~G.; and Krause, A., eds., \emph{Proceedings of the 35th
  International Conference on Machine Learning, {ICML} 2018,
  Stockholmsm{\"{a}}ssan, Stockholm, Sweden, July 10-15, 2018}, volume~80 of
  \emph{Proceedings of Machine Learning Research}, 284--293. {PMLR}.
\newblock \urlprefix\url{http://proceedings.mlr.press/v80/athalye18b.html}.

\bibitem[{Biswas and Milanfar(2017)}]{DBLP:journals/tip/BiswasM17}
Biswas, S.~K.; and Milanfar, P. 2017.
\newblock Linear Support Tensor Machine With {LSK} Channels: Pedestrian
  Detection in Thermal Infrared Images.
\newblock \emph{{IEEE} Trans. Image Process.} 26(9): 4229--4242.
\newblock \doi{10.1109/TIP.2017.2705426}.
\newblock \urlprefix\url{https://doi.org/10.1109/TIP.2017.2705426}.

\bibitem[{Cai and Vasconcelos(2018)}]{DBLP:conf/cvpr/CaiV18}
Cai, Z.; and Vasconcelos, N. 2018.
\newblock Cascade {R-CNN:} Delving Into High Quality Object Detection.
\newblock In \emph{2018 {IEEE} Conference on Computer Vision and Pattern
  Recognition, {CVPR} 2018, Salt Lake City, UT, USA, June 18-22, 2018},
  6154--6162. {IEEE} Computer Society.
\newblock \doi{10.1109/CVPR.2018.00644}.
\newblock
  \urlprefix\url{https://openaccess.thecvf.com/content_cvpr_2018/papers/Cai_Cascade_R-CNN_Delving_CVPR_2018_paper.pdf}.

\bibitem[{Carlini and Wagner(2017)}]{DBLP:conf/sp/Carlini017}
Carlini, N.; and Wagner, D.~A. 2017.
\newblock Towards Evaluating the Robustness of Neural Networks.
\newblock In \emph{2017 {IEEE} Symposium on Security and Privacy, {SP} 2017,
  San Jose, CA, USA, May 22-26, 2017}, 39--57. {IEEE} Computer Society.
\newblock \doi{10.1109/SP.2017.49}.
\newblock \urlprefix\url{https://doi.org/10.1109/SP.2017.49}.

\bibitem[{Dong et~al.(2018)Dong, Liao, Pang, Su, Zhu, Hu, and
  Li}]{DBLP:conf/cvpr/DongLPS0HL18}
Dong, Y.; Liao, F.; Pang, T.; Su, H.; Zhu, J.; Hu, X.; and Li, J. 2018.
\newblock Boosting Adversarial Attacks With Momentum.
\newblock In \emph{2018 {IEEE} Conference on Computer Vision and Pattern
  Recognition, {CVPR} 2018, Salt Lake City, UT, USA, June 18-22, 2018},
  9185--9193. {IEEE} Computer Society.
\newblock \doi{10.1109/CVPR.2018.00957}.
\newblock \urlprefix\url{http://arxiv.org/abs/1710.06081}.

\bibitem[{Eykholt et~al.(2018)Eykholt, Evtimov, Fernandes, Li, Rahmati, Xiao,
  Prakash, Kohno, and Song}]{DBLP:conf/cvpr/EykholtEF0RXPKS18}
Eykholt, K.; Evtimov, I.; Fernandes, E.; Li, B.; Rahmati, A.; Xiao, C.;
  Prakash, A.; Kohno, T.; and Song, D. 2018.
\newblock Robust Physical-World Attacks on Deep Learning Visual Classification.
\newblock In \emph{2018 {IEEE} Conference on Computer Vision and Pattern
  Recognition, {CVPR} 2018, Salt Lake City, UT, USA, June 18-22, 2018},
  1625--1634. {IEEE} Computer Society.
\newblock \doi{10.1109/CVPR.2018.00175}.
\newblock \urlprefix\url{http://arxiv.org/abs/1707.08945}.

\bibitem[{Goodfellow, Shlens, and
  Szegedy(2015)}]{DBLP:journals/corr/GoodfellowSS14}
Goodfellow, I.~J.; Shlens, J.; and Szegedy, C. 2015.
\newblock Explaining and Harnessing Adversarial Examples.
\newblock In Bengio, Y.; and LeCun, Y., eds., \emph{3rd International
  Conference on Learning Representations, {ICLR} 2015, San Diego, CA, USA, May
  7-9, 2015, Conference Track Proceedings}.
\newblock \urlprefix\url{http://arxiv.org/abs/1412.6572}.

\bibitem[{Jiang et~al.(2020)Jiang, Dong, Chen, and
  Xu}]{DBLP:journals/access/JiangDCX20}
Jiang, Y.; Dong, L.; Chen, Y.; and Xu, W. 2020.
\newblock An Infrared Small Target Detection Algorithm Based on Peak
  Aggregation and Gaussian Discrimination.
\newblock \emph{{IEEE} Access} 8: 106214--106225.
\newblock \doi{10.1109/ACCESS.2020.3000227}.
\newblock \urlprefix\url{https://doi.org/10.1109/ACCESS.2020.3000227}.

\bibitem[{Karpathy et~al.(2014)Karpathy, Toderici, Shetty, Leung, Sukthankar,
  and Li}]{DBLP:conf/cvpr/KarpathyTSLSF14}
Karpathy, A.; Toderici, G.; Shetty, S.; Leung, T.; Sukthankar, R.; and Li, F.
  2014.
\newblock Large-Scale Video Classification with Convolutional Neural Networks.
\newblock In \emph{2014 {IEEE} Conference on Computer Vision and Pattern
  Recognition, {CVPR} 2014, Columbus, OH, USA, June 23-28, 2014}, 1725--1732.
  {IEEE} Computer Society.
\newblock \doi{10.1109/CVPR.2014.223}.
\newblock \urlprefix\url{https://doi.org/10.1109/CVPR.2014.223}.

\bibitem[{Kristo, Ivasic{-}Kos, and
  Pobar(2020)}]{DBLP:journals/access/KristoIP20}
Kristo, M.; Ivasic{-}Kos, M.; and Pobar, M. 2020.
\newblock Thermal Object Detection in Difficult Weather Conditions Using
  {YOLO}.
\newblock \emph{{IEEE} Access} 8: 125459--125476.
\newblock \doi{10.1109/ACCESS.2020.3007481}.
\newblock \urlprefix\url{https://doi.org/10.1109/ACCESS.2020.3007481}.

\bibitem[{Kurakin, Goodfellow, and Bengio(2017)}]{DBLP:conf/iclr/KurakinGB17}
Kurakin, A.; Goodfellow, I.~J.; and Bengio, S. 2017.
\newblock Adversarial Machine Learning at Scale.
\newblock In \emph{5th International Conference on Learning Representations,
  {ICLR} 2017, Toulon, France, April 24-26, 2017, Conference Track
  Proceedings}.
\newblock \urlprefix\url{https://openreview.net/forum?id=BJm4T4Kgx}.

\bibitem[{Lin et~al.(2017)Lin, Goyal, Girshick, He, and
  Doll{\'{a}}r}]{DBLP:conf/iccv/LinGGHD17}
Lin, T.; Goyal, P.; Girshick, R.~B.; He, K.; and Doll{\'{a}}r, P. 2017.
\newblock Focal Loss for Dense Object Detection.
\newblock In \emph{{IEEE} International Conference on Computer Vision, {ICCV}
  2017, Venice, Italy, October 22-29, 2017}, 2999--3007. {IEEE} Computer
  Society.
\newblock \doi{10.1109/ICCV.2017.324}.
\newblock \urlprefix\url{https://doi.org/10.1109/ICCV.2017.324}.

\bibitem[{Liu et~al.(2019)Liu, Liu, Fan, Ma, Zhang, Xie, and
  Tao}]{DBLP:conf/aaai/LiuLFMZXT19}
Liu, A.; Liu, X.; Fan, J.; Ma, Y.; Zhang, A.; Xie, H.; and Tao, D. 2019.
\newblock Perceptual-Sensitive {GAN} for Generating Adversarial Patches.
\newblock In \emph{The Thirty-Third {AAAI} Conference on Artificial
  Intelligence, {AAAI} 2019, The Thirty-First Innovative Applications of
  Artificial Intelligence Conference, {IAAI} 2019, The Ninth {AAAI} Symposium
  on Educational Advances in Artificial Intelligence, {EAAI} 2019, Honolulu,
  Hawaii, USA, January 27 - February 1, 2019}, 1028--1035. {AAAI} Press.
\newblock \doi{10.1609/aaai.v33i01.33011028}.
\newblock \urlprefix\url{https://doi.org/10.1609/aaai.v33i01.33011028}.

\bibitem[{Madry et~al.(2018)Madry, Makelov, Schmidt, Tsipras, and
  Vladu}]{DBLP:conf/iclr/MadryMSTV18}
Madry, A.; Makelov, A.; Schmidt, L.; Tsipras, D.; and Vladu, A. 2018.
\newblock Towards Deep Learning Models Resistant to Adversarial Attacks.
\newblock In \emph{6th International Conference on Learning Representations,
  {ICLR} 2018, Vancouver, BC, Canada, April 30 - May 3, 2018, Conference Track
  Proceedings}.
\newblock \urlprefix\url{https://openreview.net/forum?id=rJzIBfZAb}.

\bibitem[{Moosavi{-}Dezfooli et~al.(2017)Moosavi{-}Dezfooli, Fawzi, Fawzi, and
  Frossard}]{DBLP:conf/cvpr/Moosavi-Dezfooli17}
Moosavi{-}Dezfooli, S.; Fawzi, A.; Fawzi, O.; and Frossard, P. 2017.
\newblock Universal Adversarial Perturbations.
\newblock In \emph{2017 {IEEE} Conference on Computer Vision and Pattern
  Recognition, {CVPR} 2017, Honolulu, HI, USA, July 21-26, 2017}, 86--94.
  {IEEE} Computer Society.
\newblock \doi{10.1109/CVPR.2017.17}.
\newblock \urlprefix\url{https://doi.org/10.1109/CVPR.2017.17}.

\bibitem[{Redmon and Farhadi(2017)}]{DBLP:conf/cvpr/RedmonF17}
Redmon, J.; and Farhadi, A. 2017.
\newblock {YOLO9000:} Better, Faster, Stronger.
\newblock In \emph{2017 {IEEE} Conference on Computer Vision and Pattern
  Recognition, {CVPR} 2017, Honolulu, HI, USA, July 21-26, 2017}, 6517--6525.
  {IEEE} Computer Society.
\newblock \doi{10.1109/CVPR.2017.690}.
\newblock \urlprefix\url{https://doi.org/10.1109/CVPR.2017.690}.

\bibitem[{Redmon and Farhadi(2018)}]{DBLP:journals/corr/abs-1804-02767}
Redmon, J.; and Farhadi, A. 2018.
\newblock YOLOv3: An Incremental Improvement.
\newblock \emph{CoRR} abs/1804.02767.
\newblock \urlprefix\url{http://arxiv.org/abs/1804.02767}.

\bibitem[{Ren et~al.(2017)Ren, He, Girshick, and
  Sun}]{DBLP:journals/pami/RenHG017}
Ren, S.; He, K.; Girshick, R.~B.; and Sun, J. 2017.
\newblock Faster {R-CNN:} Towards Real-Time Object Detection with Region
  Proposal Networks.
\newblock \emph{{IEEE} Trans. Pattern Anal. Mach. Intell.} 39(6): 1137--1149.
\newblock \doi{10.1109/TPAMI.2016.2577031}.
\newblock \urlprefix\url{https://doi.org/10.1109/TPAMI.2016.2577031}.

\bibitem[{Sharif et~al.(2016)Sharif, Bhagavatula, Bauer, and
  Reiter}]{DBLP:conf/ccs/SharifBBR16}
Sharif, M.; Bhagavatula, S.; Bauer, L.; and Reiter, M.~K. 2016.
\newblock Accessorize to a Crime: Real and Stealthy Attacks on State-of-the-Art
  Face Recognition.
\newblock In Weippl, E.~R.; Katzenbeisser, S.; Kruegel, C.; Myers, A.~C.; and
  Halevi, S., eds., \emph{Proceedings of the 2016 {ACM} {SIGSAC} Conference on
  Computer and Communications Security, Vienna, Austria, October 24-28, 2016},
  1528--1540. {ACM}.
\newblock \doi{10.1145/2976749.2978392}.
\newblock \urlprefix\url{https://doi.org/10.1145/2976749.2978392}.

\bibitem[{Szegedy et~al.(2014)Szegedy, Zaremba, Sutskever, Bruna, Erhan,
  Goodfellow, and Fergus}]{DBLP:journals/corr/SzegedyZSBEGF13}
Szegedy, C.; Zaremba, W.; Sutskever, I.; Bruna, J.; Erhan, D.; Goodfellow,
  I.~J.; and Fergus, R. 2014.
\newblock Intriguing properties of neural networks.
\newblock In Bengio, Y.; and LeCun, Y., eds., \emph{2nd International
  Conference on Learning Representations, {ICLR} 2014, Banff, AB, Canada, April
  14-16, 2014, Conference Track Proceedings}.
\newblock \urlprefix\url{http://arxiv.org/abs/1312.6199}.

\bibitem[{Thys, Ranst, and Goedem{\'{e}}(2019)}]{DBLP:conf/cvpr/ThysRG19}
Thys, S.; Ranst, W.~V.; and Goedem{\'{e}}, T. 2019.
\newblock Fooling Automated Surveillance Cameras: Adversarial Patches to Attack
  Person Detection.
\newblock In \emph{{IEEE} Conference on Computer Vision and Pattern Recognition
  Workshops, {CVPR} Workshops 2019, Long Beach, CA, USA, June 16-20, 2019},
  49--55. Computer Vision Foundation / {IEEE}.
\newblock \doi{10.1109/CVPRW.2019.00012}.
\newblock \urlprefix\url{http://arxiv.org/abs/1904.08653}.

\bibitem[{Xiao et~al.(2018)Xiao, Li, Zhu, He, Liu, and
  Song}]{DBLP:conf/ijcai/XiaoLZHLS18}
Xiao, C.; Li, B.; Zhu, J.; He, W.; Liu, M.; and Song, D. 2018.
\newblock Generating Adversarial Examples with Adversarial Networks.
\newblock In Lang, J., ed., \emph{Proceedings of the Twenty-Seventh
  International Joint Conference on Artificial Intelligence, {IJCAI} 2018, July
  13-19, 2018, Stockholm, Sweden}, 3905--3911. ijcai.org.
\newblock \doi{10.24963/ijcai.2018/543}.
\newblock \urlprefix\url{https://doi.org/10.24963/ijcai.2018/543}.

\bibitem[{Xu et~al.(2019)Xu, Zhang, Liu, Fan, Sun, Chen, Chen, Wang, and
  Lin}]{DBLP:journals/corr/abs-1910-11099}
Xu, K.; Zhang, G.; Liu, S.; Fan, Q.; Sun, M.; Chen, H.; Chen, P.; Wang, Y.; and
  Lin, X. 2019.
\newblock Evading Real-Time Person Detectors by Adversarial T-shirt.
\newblock \emph{CoRR} abs/1910.11099.
\newblock \urlprefix\url{http://arxiv.org/abs/1910.11099}.

\bibitem[{Yin et~al.(2020)Yin, Yiu, Hu, and
  Tang}]{DBLP:journals/Cognitive/ZiYin}
Yin, Z.; Yiu, V.; Hu, X.; and Tang, L. 2020.
\newblock End-to-End Face Parsing via Interlinked Convolutional Neural
  Networks.
\newblock \emph{Cognitive Neurodynamics}
  \doi{https://doi.org/10.1007/s11571-020-09615-4}.

\bibitem[{Zhao et~al.(2019)Zhao, Cheng, Zhou, Zhang, and
  Pan}]{DBLP:conf/apsipa/ZhaoCZZP19}
Zhao, Y.; Cheng, J.; Zhou, W.; Zhang, C.; and Pan, X. 2019.
\newblock Infrared Pedestrian Detection with Converted Temperature Map.
\newblock In \emph{2019 Asia-Pacific Signal and Information Processing
  Association Annual Summit and Conference, {APSIPA} {ASC} 2019, Lanzhou,
  China, November 18-21, 2019}, 2025--2031. {IEEE}.
\newblock \doi{10.1109/APSIPAASC47483.2019.9023228}.
\newblock \urlprefix\url{https://doi.org/10.1109/APSIPAASC47483.2019.9023228}.

\bibitem[{Zhou et~al.(2018)Zhou, Tang, Wang, Han, Liu, and
  Zhang}]{DBLP:journals/corr/abs-1803-04683}
Zhou, Z.; Tang, D.; Wang, X.; Han, W.; Liu, X.; and Zhang, K. 2018.
\newblock Invisible Mask: Practical Attacks on Face Recognition with Infrared.
\newblock \emph{CoRR} abs/1803.04683.
\newblock \urlprefix\url{http://arxiv.org/abs/1803.04683}.

\end{thebibliography}

\renewcommand{\thefigure}{S\arabic{figure}}
\setcounter{figure}{0}

\clearpage
\begin{center}
    \huge{ \textbf{Supplementary Material: Fooling thermal infrared pedestrian detectors in real world using small bulbs}}
\end{center}
\section{Details of the pixel-level adversarial patch attack in the digital world}
The pixel-level adversarial patch used pixels as the basic unit. Each pixel value was an optimization variable.
We followed the work of Thys et al. (2019) \nocite{DBLP:conf/cvpr/ThysRG19}to build a square patch with the pixel size of 300$\times$300.
The difference was that our patch was a grayscale image instead of an RGB image. We first initialized 
a 300$\times$300 pixel-level patch. There were two options: random initialization and uniform initialization. We found that when each pixel was initialized to 0.5, the network 
convergence effect was better in our experiment, so we adopted the uniform initialization method. 
% The image annotation can generate several bounding boxes, and we place patches in the relative positions of these bounding boxes.

\begin{figure}[htb]
\centering
\includegraphics[width=0.55\columnwidth]{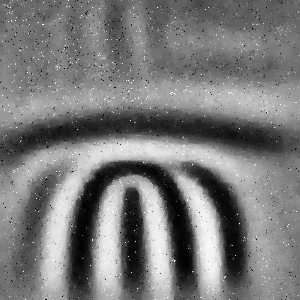} % Reduce the figure size so that it is slightly narrower than the column. Don't use precise values for figure width.This setup will avoid overfull boxes. 
\caption{An example of the pixel-level adversarial patch.}
\label{pixel-patch}
\end{figure}

To make the patch more robust, we designed a variety of transformations including random noise on the 
patch, random rotation of the patch (clockwise or counterclockwise within 20 degrees), random translation of 
the patch, and random changes in the brightness and contrast of the patch. These transformations simulate the 
perturbation of the physical world to a certain extent, which effectively improves the robustness of the patch. 
Then we used the training set of \textit{FLIR\_person\_select} and placed the patch on the upper body of the pedestrians 
according to the position of the bounding box. The size of the patch was 1/5 of the height of the bounding box. 
Next, we used the patched image as input and ran the YOLOv3 pedestrian detector we had trained. We used a 
stochastic gradient descent optimizer with momentum, and the size of each batch was 8. The optimizer used the 
back-propagation algorithm to update the pixel values by minimizing the loss function. Through this process, 
we obtained a series of patches. Figure \ref{pixel-patch} is an example after 65801 iterations.

\begin{figure}[htb]
\centering
\includegraphics[width=0.9\columnwidth]{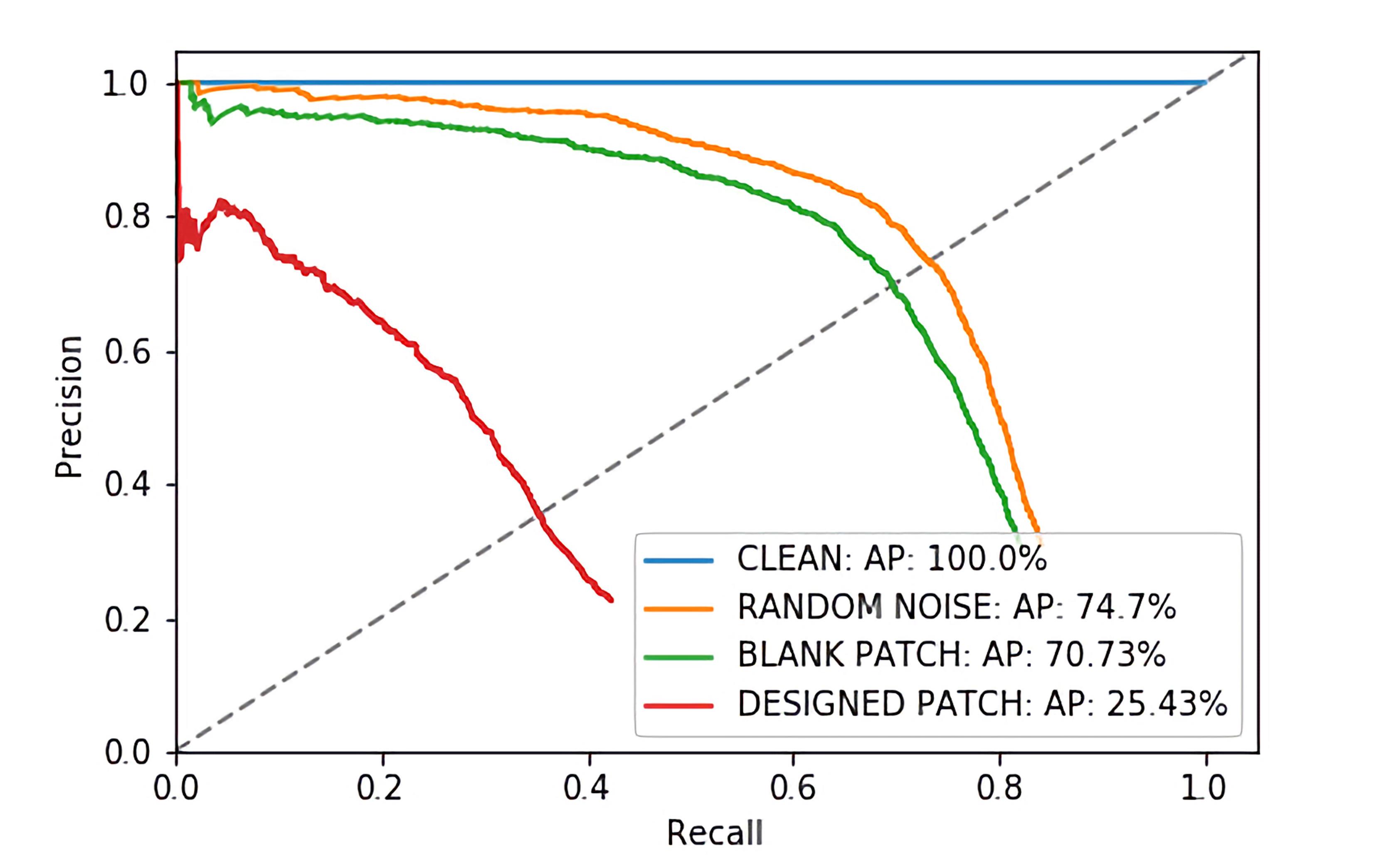} % Reduce the figure size so that it is slightly narrower than the column. Don't use precise values for figure width.This setup will avoid overfull boxes. 
\caption{Evaluation of the pixel-level adversarial patch attack.}
\label{pixel-patch-evaluation}
\end{figure}

\begin{figure}[htb]
\centering
\includegraphics[width=0.9\columnwidth]{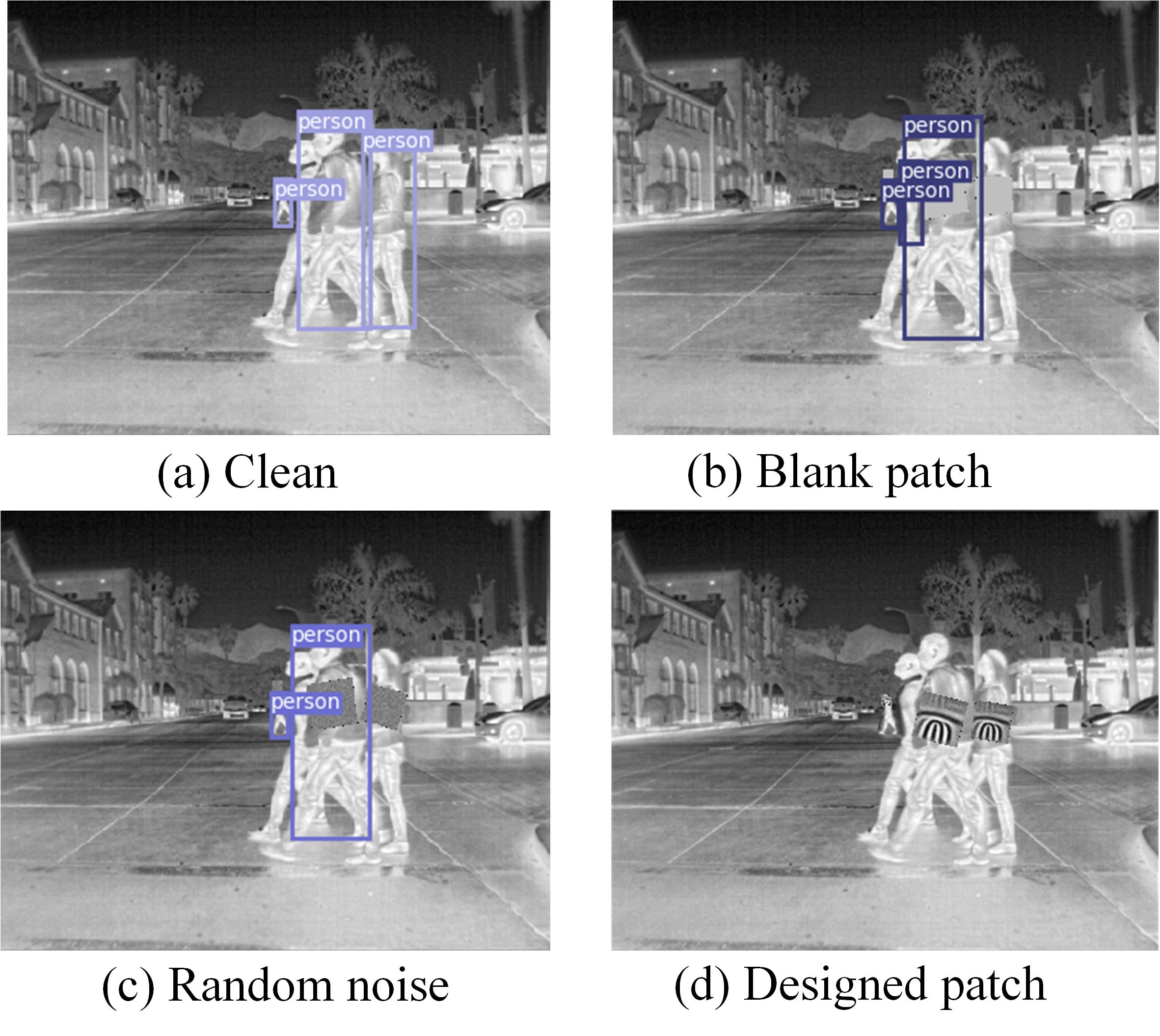} % Reduce the figure size so that it is slightly narrower than the column. Don't use precise values for figure width.This setup will avoid overfull boxes. 
\caption{The pixel-level adversarial patch attack and control experiments.}
\label{pixel-examples}
\end{figure}

Next, we applied the optimized patch shown in Figure \ref{pixel-patch} to the test set, using the same process we 
used during training, including various transformations. We used random noise patches with maximum amplitude 
value 1 and constant pixel value patches (blank patches) for control experiments. The pixel values of blank 
patch in our experiment were 0.75 
(Other values had similar influence to detectors; see the next section).
% We tried other values and showed the attacking performance of blank patches with different pixel values in the section \textit{The effects of pixel-value of blank patch}. 
We applied these different patches to the \textit{FLIR\_person\_select} test set, and 
then input the patched images to the same detection network to test its detection performance. We adopted the 
IOU method to calculate the accuracy of the detection. The precision-recall (PR) curves are shown in Figure \ref{pixel-patch-evaluation}. 
We defined the output of the clean image input as the ground truth, then the pixel-level adversarial patch made the average 
precision (AP, the area under the PR curve) of the target detector drop by 74.57\%. An example is shown in Figure \ref{pixel-examples}. 
In contrast, the AP of the target detector dropped by 25.30\% and 29.27\% 
using random noise patch and blank patch, respectively. 

\section{Influence of the pixel value of the blank patch to the detection performance}
\begin{figure}[htb]
\centering
\includegraphics[width=0.82\columnwidth]{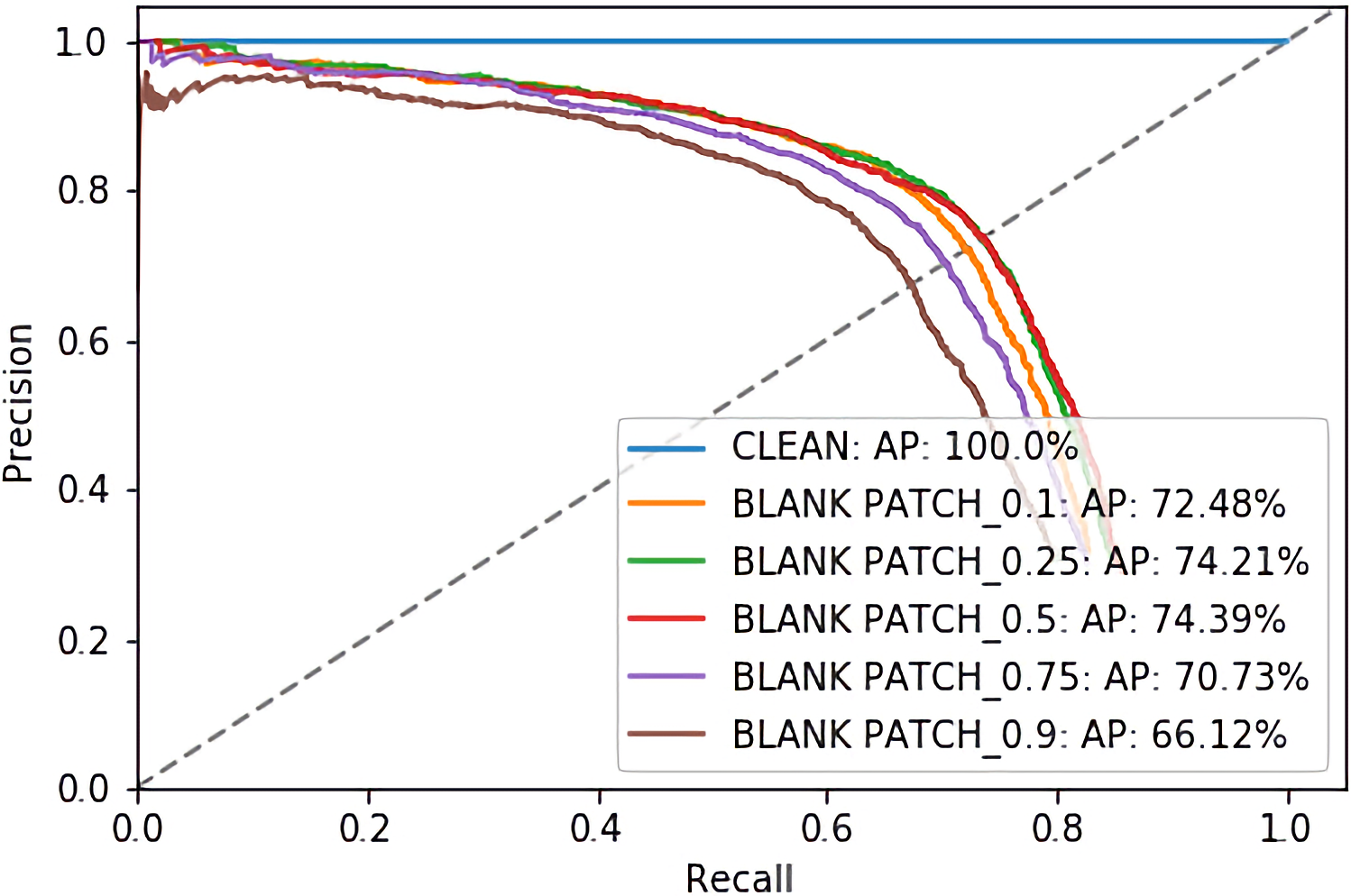} % Reduce the figure size so that it is slightly narrower than the column. Don't use precise values for figure width.This setup will avoid overfull boxes. 
\caption{Evaluation of blank patches with different pixel values.}
\label{hb_values}
\end{figure}

All pixels of the blank patch had the same value. So we studied the influence of the pixel value to the 
detection performance. The pixel value varied from 0 to 1. We chose five values (0.1, 0.25, 
0.5, 0.75 and 0.9). The PR curves are shown in Figure \ref{hb_values}. The blank patches with different pixel values caused the 
AP of YOLOv3 to drop by 30\% $\pm$ 5\%. 
% The attacking performance of blank patches with different pixel values did not vary too much. 
Therefore the influence to the detector did not vary significantly with different pixel values. 
We chose a typical value of 0.75 in other experiments. 

\section{Attack the visible light and infrared object detection systems at the same time}
\begin{figure}[htb]
\centering
\includegraphics[width=0.93\columnwidth]{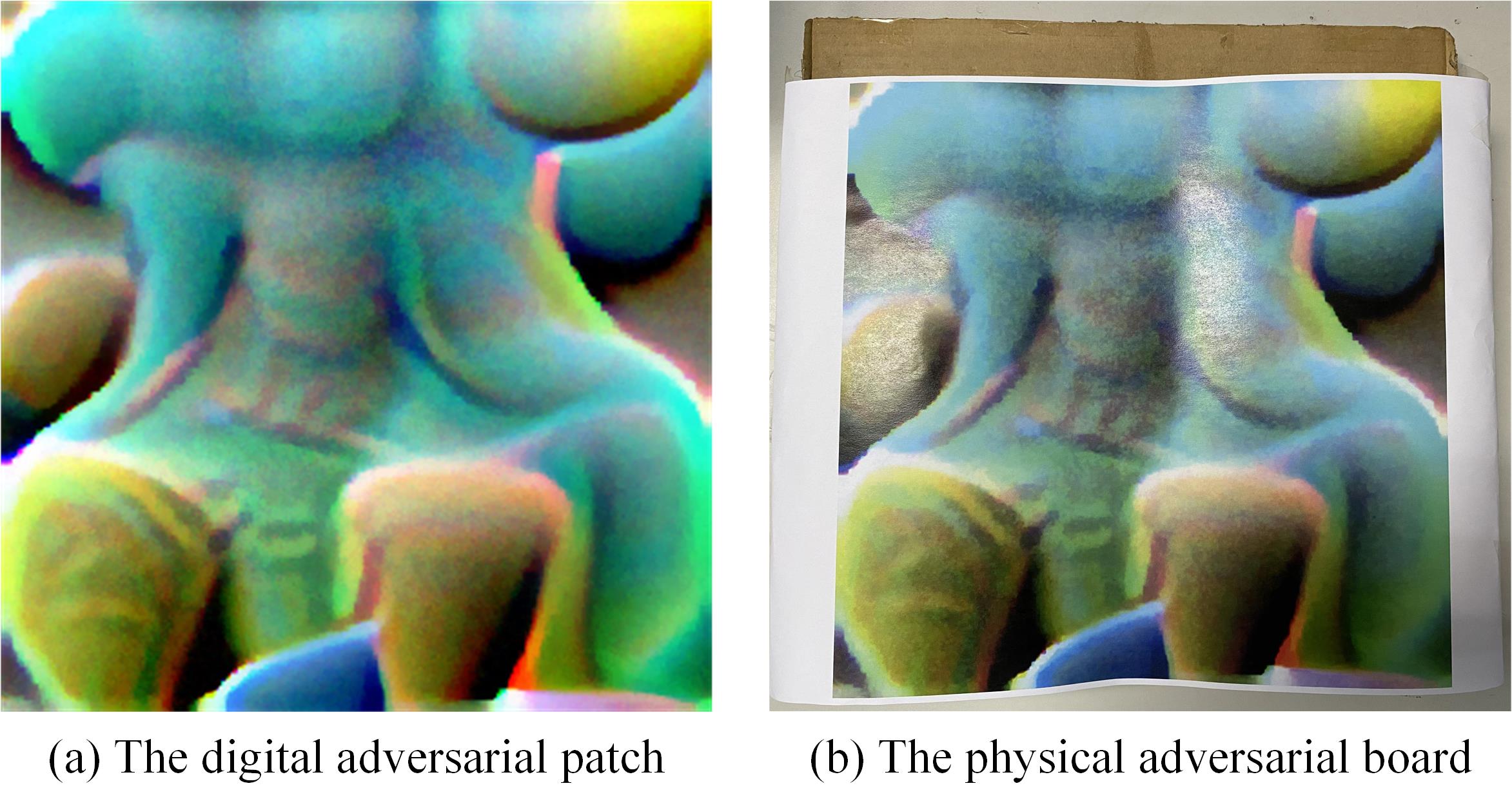} % Reduce the figure size so that it is slightly narrower than the column. Don't use precise values for figure width.This setup will avoid overfull boxes. 
\caption{The adversarial digital patch and physical board. Note that the small bulbs are covered by the printed paper in (b).}
\label{phy-dig}
\end{figure}

% Thermal infrared imaging systems can detect objects behind certain obstacles. At first, we followed the work of 
% Thys et al. (2019) to print an adversarial patch on a A3 paper, which could attack YOLOv3 successfully in the 
% visible light images. 
An interesting question is whether we can design a physical board that can evade the person detectors working on 
both visible light images and infrared images. Based on our method, the solution turned out to be simple. We 
printed an adversarial patch on a paper, which was crafted according to the a previous work \cite{DBLP:conf/cvpr/ThysRG19} by using 
YOLOv3 as the target detector. 
The size of the adversarial patch was 29.8cm $\times$ 29.8cm. Then we put the paper on the physical board with small bulbs 
% which could attack thermal infrared object detection systems. 
we designed before.
The digital patch and the physical board is shown in Figure \ref{phy-dig}(a) and Figure \ref{phy-dig}(b). 

\begin{figure}[htb]
\centering
\includegraphics[width=0.8\columnwidth]{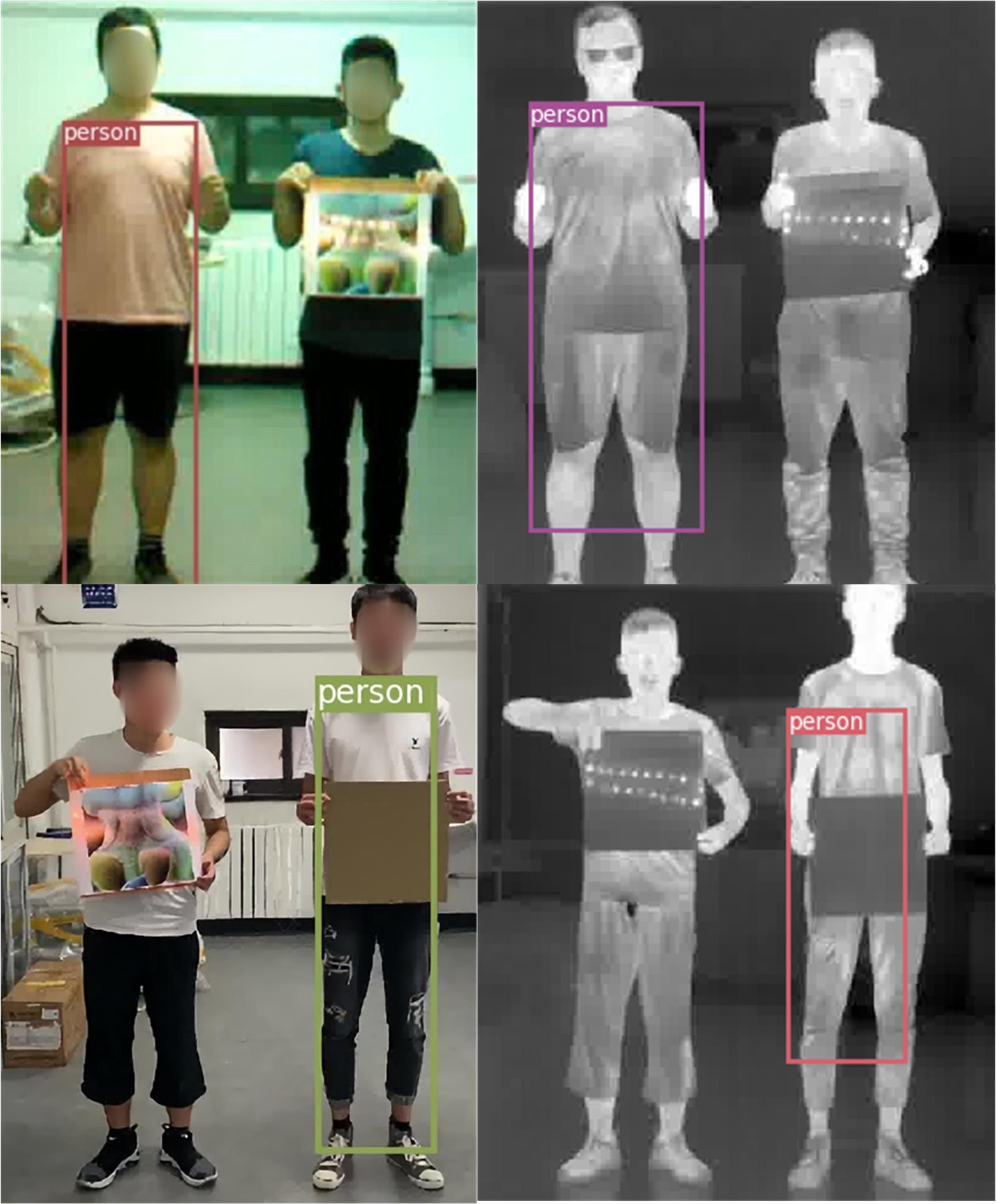} % Reduce the figure size so that it is slightly narrower than the column. Don't use precise values for figure width.This setup will avoid overfull boxes. 
\caption{An example of visible light and infrared physical board attacks. For privacy reasons, we blurred the facial area on visible light images.}
\label{VIS-TIR}
\end{figure}

% After that we invited several people, 
We invited several persons to participate in the test. 
They could hold the adversarial board, or a blank board, or nothing. We used visible light camera and thermal 
infrared camera to shoot these people under the same conditions. Then we input the images to the target detector 
YOLOv3. The result showed that we could successfully attack the visible light and infrared object detection systems at the same 
time. Several examples are shown in Figure \ref{VIS-TIR}. To the best of our knowledge, this is the first multispectral (infrared and visible) attack in the physical world.

\end{document}